\documentclass[10pt,twocolumn,letterpaper]{article}
\usepackage{mytitlesec}

\usepackage{multirow}
\usepackage[pagenumbers]{cvpr} 

\usepackage{cuted}
\usepackage{bbding}
\usepackage{pifont}

\definecolor{cvprblue}{rgb}{0.21,0.49,0.74}
\usepackage[pagebackref,breaklinks,colorlinks,allcolors=cvprblue]{hyperref}

\def\confName{CVPR}
\def\confYear{2025}

\title{Bench2Drive-R: Turning Real World Data into Reactive Closed-Loop Autonomous Driving Benchmark by Generative Model}

\author{Junqi You*
\and Xiaosong Jia* \and Zhiyuan Zhang \and Yutao Zhu \and  Junchi Yan$^{\dagger}$ \\ \\
  \normalsize{$^*$ Equal contributions.  \quad $^\dagger$ Correspondence author} \\ \\
  Dept. of CSE \& School of AI \& MoE Key Lab of AI, Shanghai Jiao Tong University\\ \\ 
}

\begin{document}
\maketitle

\begin{abstract}
\vspace{-4mm}

For end-to-end autonomous driving (E2E-AD), the evaluation system remains an open problem. Existing closed-loop evaluation protocols usually rely on simulators like CARLA being less realistic; while NAVSIM using real-world vision data, yet is limited to fixed planning trajectories in short horizon and assumes other agents are not reactive. 

We introduce \textbf{Bench2Drive-R}, a generative framework that enables reactive closed-loop evaluation. Unlike existing video generative models for AD, the proposed designs are tailored for interactive simulation, where sensor rendering and behavior rollout are decoupled by applying a separate behavioral controller to simulate the reactions of surrounding agents. As a result, the renderer could focus on image fidelity, control adherence, and spatial-temporal coherence. For temporal consistency, due to the step-wise interaction nature of simulation, we design a noise modulating temporal encoder with Gaussian blurring to encourage long-horizon autoregressive rollout of image sequences without deteriorating distribution shifts. For spatial consistency, a retrieval mechanism, which takes the spatially nearest images as references, is introduced to to ensure scene-level rendering fidelity during the generation process. The spatial relations between target and reference are explicitly modeled with 3D relative position encodings and the potential over-reliance of reference images is mitigated with hierarchical sampling and classifier-free guidance.

We compare the generation quality of Bench2Drive-R with existing generative models and achieve state-of-the-art performance. We further integrate Bench2Drive-R into nuPlan and evaluate the generative qualities with closed-loop simulation results. We will open source our code.
\end{abstract}
\vspace{-5mm}

\begin{table}
    \centering
    \caption{\textbf{Evaluation Framework for E2E-AD models}. } 
    \resizebox{\columnwidth}{!}{
    \begin{tabular}{c|cccc}
    \hline \toprule
      & Sensor & Human & Ego & \multirow{2}{*}{Reactive} \\
      & Fiedlity & Behavior & Movement & \\
    \midrule
    Open-Loop~\cite{nuscenes}& \textcolor{teal}{\ding{51}} & \textcolor{teal}{\ding{51}} &  \textcolor{red}{\ding{55}}& \textcolor{red}{\ding{55}} \\ Simulator~\cite{dosovitskiy2017carla,jia2024b2d} & \textcolor{red}{\ding{55}} & \textcolor{red}{\ding{55}} &	\textcolor{teal}{\ding{51}} & \textcolor{teal}{\ding{51}}\\ NAVSIM~\cite{dauner2024navsim}&	  \textcolor{teal}{\ding{51}} & \textcolor{teal}{\ding{51}} & \textcolor{teal}{\ding{51}} & \textcolor{red}{\ding{55}} \\
    \textbf{Bench2Drive-R} & \textcolor{teal}{\ding{51}} & \textcolor{teal}{\ding{51}} &	\textcolor{teal}{\ding{51}}&	\textcolor{teal}{\ding{51}} 
    \\ \bottomrule \hline
    \end{tabular}
    }\label{tab:e2e_eval}
\end{table}

\begin{figure*}
    \centering
\includegraphics[width=1.0\textwidth]{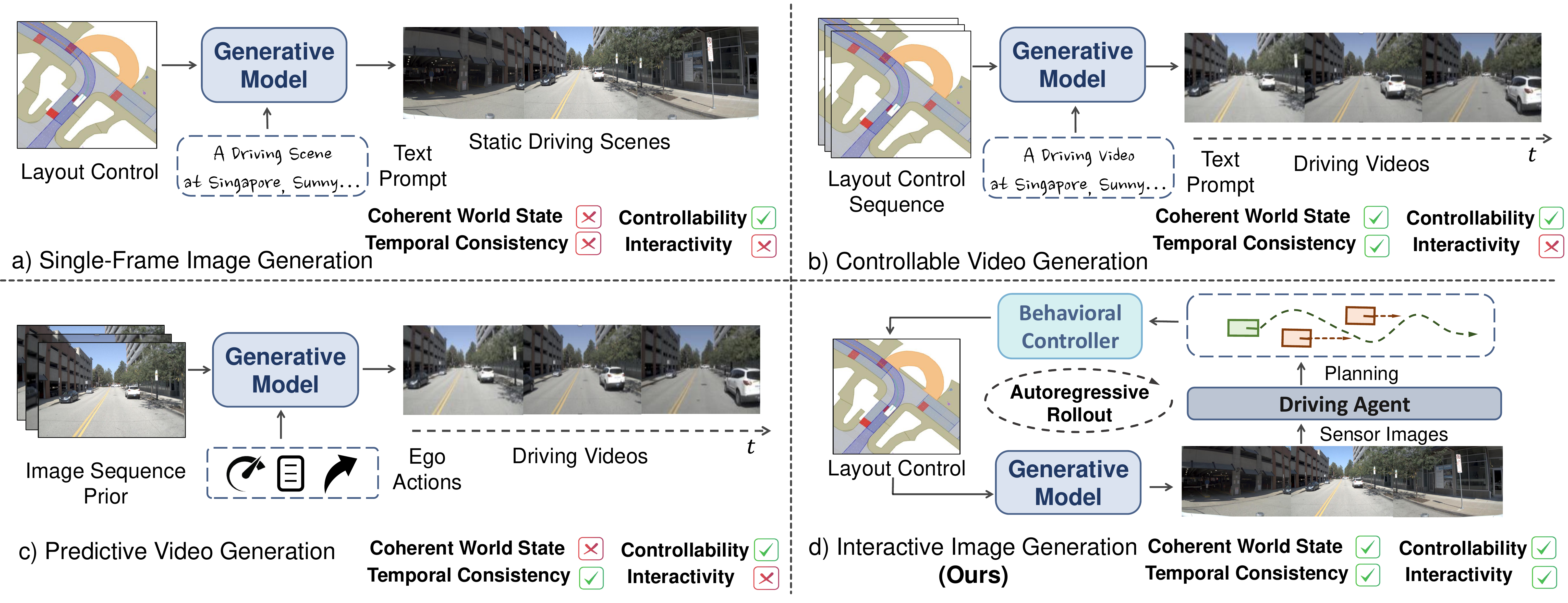}
\captionof{figure}{\textbf{Different Paradigms of Generative Models for Autonomous Driving}: (a) \textbf{Single-Frame Image Generation}~\cite{swerdlow2024bevgen,yang2023bevcontrol,gao2024magicdrive}, as relatively early works, do not account for temporal generation.  (b) \textbf{Controllable Video Generation}~\cite{wen2024panaceaplus, ma2024delphi, wang2023drivedreamer} focuses on generating videos with controls for each frame, which is not suitable for interactive simulation. (c) \textbf{Predictive Video Generation}~\cite{gao2024vista, wang2023drivingfuturemultiviewvisual, hu2023gaia} emphasizes the annotation-free training ability, which lacks the ability to adhere to control.  (d) \textbf{Interactive Image Generation}: The proposed framework leverages the power of generative models in an autoregressive manner, enabling high-frequency interactions with end-to-end driving models and generating temporally consistent images. The integration of a rule-based behavioral controller simulates the behavior of other driving agents to ensure a coherent world state and provides layout controls for the generative part of the framework.}
    \label{fig:paradigm}
\end{figure*}

\section{Introduction}

End-to-end autonomous driving (E2E-AD)~\cite{hu2023planning, jiang2023vad, Weng2024paradrive} has recently gained pervasive attention from both industry and academia~\cite{yang2023llm4drive}. Unlike traditional modular AD systems, E2E-AD aims to directly predict future trajectories based on raw sensor data. As still in its early stage, benchmarking E2E-AD models remains an open problem. Existing evaluation approaches can be categorized into three classes. (1) \textbf{Open-Loop Evaluation}~\cite{nuscenes} typically measures displacement errors between predicted trajectories and logged expert trajectories. Recent studies~\cite{zhai2023rethinking, li2024egostatusneedopenloop, dauner2023partingmisconceptionslearningbasedvehicle} reveal that it suffers from imbalanced datasets, heavy reliance on expert ego state, and distribution shift, etc. (2) \textbf{Closed-Loop Simulation}~\cite{Prakash2021CVPR,Chitta2023PAMI,jia2024b2d} utilizes simulators, such as CARLA~\cite{dosovitskiy2017carla}, to evaluate the planning performance in a reactive way. However, there still remain notable gaps between simulator and real world from both rendering and behavioral perspectives.  (3) Recent work \textbf{NAVSIM}~\cite{dauner2024navsim} proposes an intermediate approach situated between open-loop and closed-loop evaluation. It proposes a non-reactive simulator and collects closed-loop evaluation metrics over a short period of simulation horizon. Ego trajectories are predicted at the initial frame and kept fixed during the simulation period. While it can provide realistic sensor images and align more closely with closed-loop metrics than open-loop displacement errors, it fails to capture the model’s planning capabilities in scenarios where agent interactions~\cite{ide-net,hdgt} critically impact planning outcomes, such as merging into traffic streams or lane changes in dense traffic. We compare different paradigms for E2E-AD evaluation in Table~\ref{tab:e2e_eval}.

One promising way to address the aforementioned challenges is to develop a \textbf{reactive closed-loop simulation} framework capable of delivering authentic sensor data with high fidelity and consistency. Considerable works have explored the application of generative models to autonomous driving e.g. generating realistic sensor data, most of which have focused on generating novel driving scenes primarily for data augmentation in perception tasks. These efforts include generating static BEV-conditioned driving scenarios for detection and online mapping~\cite{gao2024magicdrive, yang2023bevcontrol, swerdlow2024bevgen}, producing video clips~\cite{wen2023panacea, wen2024panaceaplus, ma2024delphi} for tasks such as tracking and trajectory prediction, as well as using video diffusion model as world models to implicitly simulate driving scenarios~\cite{gao2024vista, wang2023drivedreamer, zhao2024drivedreamer2}. However, single-frame image generation lacks the constraint of temporal consistency while video generation is not able to conduct step-by-step interactions with E2E-AD models and thus could not be used in the closed-loop interactive simulation. Fig.~\ref{fig:paradigm} gives comparison of aforementioned  generation paradigms.

To this end, we propose \textbf{Bench2Drive-R}, an interactive generative method with autoregressive rollout for closed-loop reactive evaluation for E2E-AD models.  Specifically, Bench2Drive-R consists of two parts: \textbf{a reactive behavioral controller and a generative renderer}. We base our behavioral controller on the widely used planning benchmark -nuPlan~\cite{karnchanachari2024nuplan} while the generative renderer is based on diffusion models~\cite{ho2020ddpm} with condition controls~\cite{zhang2023controlnet}. At each iteration, E2E-AD models output planned actions based on current sensor data. Then, the reactive behavioral controller takes in this planning results and rollouts driving scenario in the bounding box level (forwarding ego state and other driving agents' states). Finally, with the new states of environments and previous steps as conditions, the generative renderer produces corresponding sensor data at the updated frame, which is fed into E2E-AD models for next action.

\begin{figure*}[!tb]
    \centering
\includegraphics[width=1.0\linewidth]{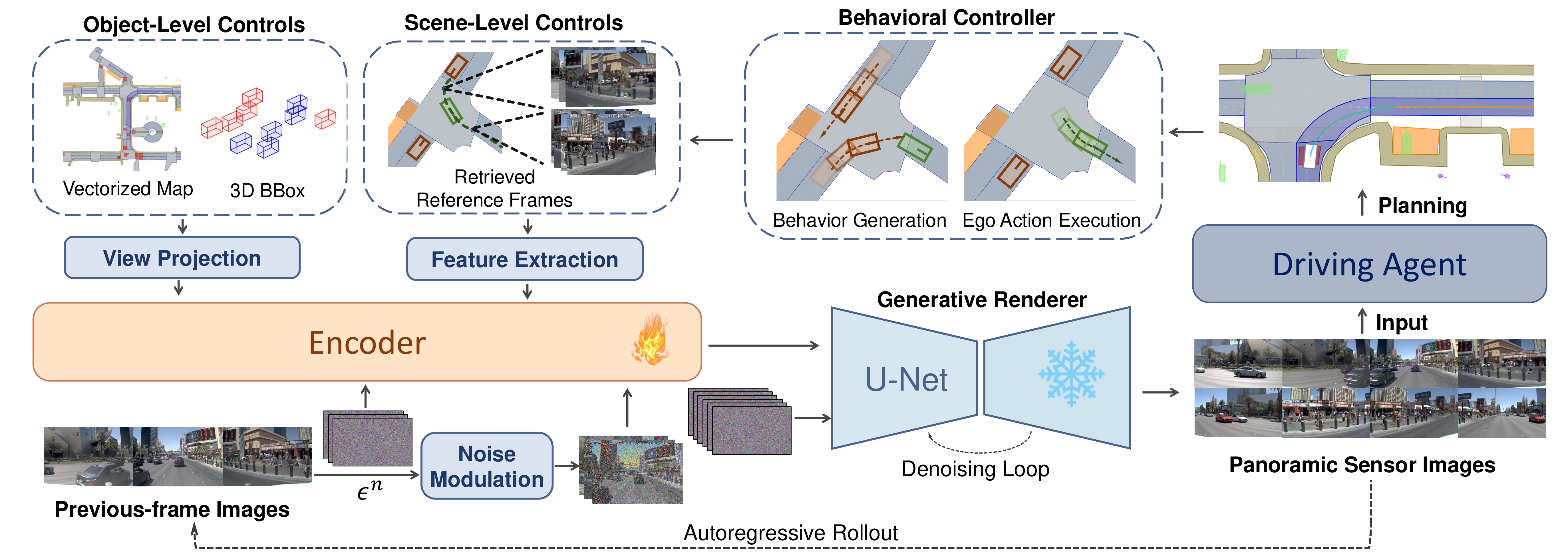}
    \caption{\textbf{Overall Framework}: The proposed Bench2Drive-R is composed of two parts: a \textbf{behavioral controller} that executes ego actions and generates behaviors of other driving agents; a \textbf{generative renderer} that produces multi-view sensor images in an autoregressive manner. To improve fidelity, the generative renderer (1) utilizes previous-frame image for temporal consistency; (2) retrieves spatially nearest reference image pair for background prior; (3) adheres to projected layout element controls for object-level consistency. 
    }
    \label{fig:structure}
\end{figure*}

As a \textbf{simulation-oriented generative renderer}, several unique characteristics could be utilized for better fidelity and consistency: (1) For \textbf{temporal consistency}, the previous frame could always be used as conditions to provide priors. However, it introduces significant train-val gaps which collapse generation easily due to the cumulative errors of autoregressive generation. Thus, we propose a noise modulation module with Gaussian blurring during the training process to let the model adapt to the defective conditions. (2) For \textbf{spatial consistency}, our key observation is that \textbf{the static background could be retrieved from the database and thus the uncertainty is eliminated.} This point is quite different from video generation models since their purpose is to provide diverse samples while our goal is to provide high-fidelity simulation. Thus, we retrieve the two frames with lowest distance in the forward and backward direction respectively and use them as conditions to guide the generation of static background.

We compare the proposed simulation-oriented generative renderer in  Bench2Drive-R with state-of-the-art generative models for AD and demonstrate superior fidelity. Further, we implement a closed-loop interactive end-to-end simulation platform based on nuPlan. We compare Bench2Drive-R  with baseline generative methods by comparing the performance of the same E2E-AD models under different renderers and Bench2Drive-R demonstrates explicit advantages. We conduct ablation studies and case analysis to evaluate the effectiveness of proposed module. 

In summary, our contributions are as follows:

\begin{itemize}
\item  We introduce \textbf{Bench2Drive-R}, the first generative (especially conditioned on real-world driving data), closed-loop reactive simulation framework serving as an end-to-end extension of the  planning-only nuPlan.
 
\item Different from the recent efforts~\cite{gao2024vista, zhao2024drivedreamer2, hu2023gaia, jia2023adriverigeneralworldmodel, wang2023drivingfuturemultiviewvisual, wang2023drivedreamer, kim2021drivegancontrollablehighqualityneural, gao2024vista, zhao2024drivedreamer2, hu2023gaia, jia2023adriverigeneralworldmodel,wang2023drivingfuturemultiviewvisual, wang2023drivedreamer, kim2021drivegancontrollablehighqualityneural, li2023drivingdiffusion, delphi, wen2024panaceaplus, wen2023panacea, huang2024subjectdrive, wu2024drivescape} of video generation, we propose several simulation-oriented designs, which significantly enhance the generation fidelity.
\item  Extensive experiments including ablation studies on the nuScenes and nuPlan dataset show the effectiveness of the whole proposed framework as well as each module.
\end{itemize}

\section{Related Works}
\subsection{Benchmarks for End-to-End Driving Models}
E2E-AD methods transform the entire AD system into a learnable network to directly optimize the planning performance. Over the years, E2E-AD has gradually developed from taking only simulated sensor data~\cite{pomerleau1988alvinn, codevilla2018cil, codevilla2019exploring, Renz2022CORL, chen2020learning, zhang2021roach, wu2022trajectoryguided, hu2022model, Prakash2021CVPR, Chitta2022PAMI, chen2022lav,shao2022interfuser, zhang2022mmfn, jia2023thinktwice, shao2023reasonnet, Jaeger2023ICCV} to training on real-world collected data~\cite{hu2023planning, jiang2023vad, Weng2024paradrive, zhang2024sparseadsparsequerycentricparadigm, sun2024sparsedriveendtoendautonomousdriving, su2024difsdegocentricfullysparse, hu2022stp3, li2024enhancingendtoendautonomousdriving, zhang2024bevworldmultimodalworldmodel, li2024hydramdpendtoendmultimodalplanning}.  Many existing E2E models are evaluated on the open-loop nuScenes protocol~\cite{nuscenes}, where the displacement errors between expert and predicted trajectories are used as metrics. However, open-loop evaluation has significant limitations, as pointed out in~\cite{zhai2023rethinking, li2024egostatusneedopenloop, dauner2023partingmisconceptionslearningbasedvehicle}. As for the closed-loop benchmarks, CARLA~\cite{dosovitskiy2017carla}'s simulation has large gaps compared to real world in both rendering and behavior level while Waymax~\cite{gulino2023waymax} and nuPlan~\cite{karnchanachari2024nuplan} are limited to bounding box level assessments. Recently, NAVSIM~\cite{dauner2024navsim} workarounds the problem by introducing an open-loop metric PDMS which shows better correlation to close-loop metrics than displacement errors. However, NAVSIM assumes other vehicles are not reactive and fixes the ego vehicle's behavior in the simulation period, which cannot reflect the driving performance under highly interactive scenarios.

In this work, we aim to build a closed-loop interactive end-to-end driving simulation, leveraging the recent huge advance in generative models.

\subsection{Generative and Restruction Models for AD}

Diffusion models~\cite{song2022ddim, ho2020ddpm} generate image by progressively denoising a randomly sampled Gaussian noise. Recent advances in the field has allowed diffusion models to generate photorealistic synthesis of images conditioned on various input including text prompts, images, etc.~\cite{li2023gligen, zhang2023controlnet, mou2023t2iadapter}

In the field of autonomous driving, many works try to synthesize novel street-view with generative models. A line of works focus on using generated images as data argumentation for downstream perception tasks. Some works~\cite{xie2023boxdiff, zhou2024simgen, yang2023bevcontrol,swerdlow2024bevgen, gao2024magicdrive, zhang2024perldiff} use bounding boxes and map polylines as controls to generate single-frame driving scene. Other works have advanced into the area of layout controlled video generation with high temporal consistency~\cite{li2023drivingdiffusion, delphi, wen2024panaceaplus, wen2023panacea, huang2024subjectdrive, wu2024drivescape, guo2024infinitydrivebreakingtimelimits, guo2024infinitydrivebreakingtimelimits, liang2023luciddreamerhighfidelitytextto3dgeneration, wu2024drivescapehighresolutioncontrollablemultiview, deng2024streetscapeslargescaleconsistentstreet} and explore the impact of synthesized data on detection~\cite{li2023delving,zhu2024flatfusiondelvingdetailssparse} and prediction~\cite{jia2022multi,jia2023towards,jia2024amp} tasks. 

Another line of works~\cite{gao2024vista, zhao2024drivedreamer2, hu2023gaia, jia2023adriverigeneralworldmodel, wang2023drivingfuturemultiviewvisual, wang2023drivedreamer, kim2021drivegancontrollablehighqualityneural, wang2024freevsgenerativeviewsynthesis} focuses on the predictive capability of generative models. They explore the possibility of turning video diffusion model into generalizable driving world model of a fully differentiable driving simulator, which rollouts world states in pixel space based on input actions and behaviors. However, such world models can't guarantee a coherent world state and fails to provide interactive interfaces for simulations. 

Other efforts are spent on closing simulation with reality with 3D reconstruction techniques, such as NeRF~\cite{guo2023streetsurfextendingmultiviewimplicit, irshad2023neo360neuralfields, lu2023urbanradiancefieldrepresentation, rematas2021urbanradiancefields, tancik2022blocknerfscalablelargescene, yang2023unisimneuralclosedloopsensor}
 and 3D GS~\cite{chen2024periodicvibrationgaussiandynamic, chen2024omnireomniurbanscene, huang2024textits3gaussianselfsupervisedstreetgaussians, yan2024streetgaussiansmodelingdynamic, yu2024sgdstreetviewsynthesis, zhou2024drivinggaussiancompositegaussiansplatting, zhao2024drivedreamer4dworldmodelseffective, ni2024recondreamercraftingworldmodels, zhao2024drivedreamer4dworldmodelseffective, fan2024freesimfreeviewpointcamerasimulation, tian2024drivingforwardfeedforward3dgaussian, yang2024drivingscenesynthesisfreeform}. However, reconstruction methods are not fit for simulation task because of their costly reconstructing processes and weak generalizability for out-of-distribution driving cases.

In summary, none of existing works are very suitable for simulation, which motivates us to design simulation oriented Bench2Drive-R, featuring  interactive generation. There are also some concurrent works~\cite{zhou2024hugsimrealtimephotorealisticclosedloop, lu2024infinicubeunboundedcontrollabledynamic, yan2024drivingspherebuildinghighfidelity4d, yang2024drivearenaclosedloopgenerativesimulation} probing into the area of closed-loop simulation with sensor data, but they either resort to spatially-restricted 3D reconstruction methods, which can't provide fine-grind controls, or fail to provide reactive, data-driven~\cite{wu2023PPGeo,lu2024activead} traffic simulation.


\section{Methods}

\begin{figure}[!tb]
    \centering
\includegraphics[width=1.0\linewidth]{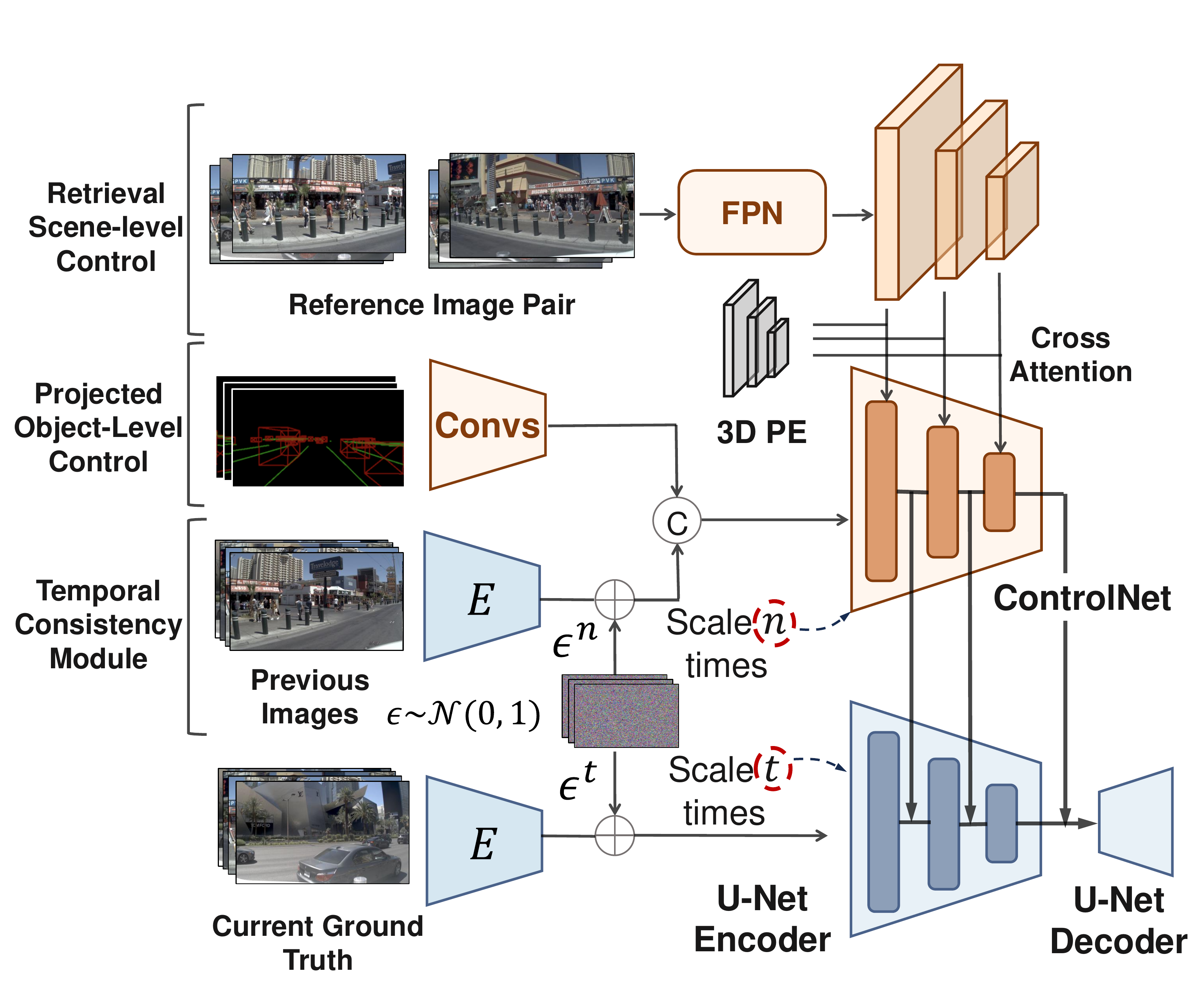}
    \caption{\textbf{ Structural Design of Generative Renderer.} We design three additive modules to ensure controllable, consistent, and interactive image generation. \textbf{a) Temporal consistency module} incorporates previous frame images. Noise modulation module helps prevent distribution drift during autoregressive rollout. \textbf{b) Projected Object-Level Control } allows fine-grind controls over the location and orientation of driving vehicles in the scenario. \textbf{c) Retrieval Scene-Level Control} ensures spatial consistency by extracting multi-level features from nearest reference image pairs injecting them into the ControlNet with attention mechanism.  \vspace{-5mm}
    }
    \label{fig:renderer}
\end{figure}

\begin{figure*}[!tb]
    \centering
\includegraphics[width=1.0\linewidth]{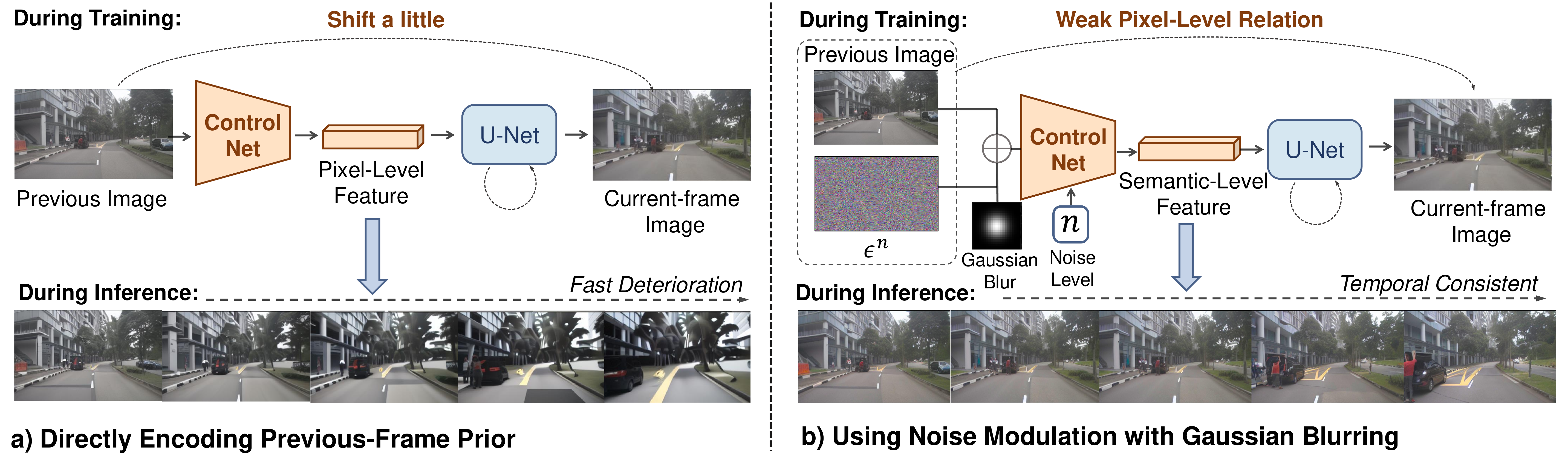}
    \caption{\textbf{Deal with Autoregressive Distribution Shift.}  \textbf{a)} During training, due to the resemblance between previous and current frame prior, the model would overly rely on previous frame. During inference, the generation errors (artifacts) will cumulate and finally collapse.
    \textbf{b)} Adding Gaussian blurring and a random level of noise to previous images can destroy obvious pixel-level relations between the two frames. Noise level $n$ is fed as inputs to give model hints on the corrupted extent. As a result, the model can adapt to degenerated previous images and learn to extract high-level prior information instead of pixel-level copy.  }
    \label{fig:prev_mech}
\end{figure*}

\subsection{Preliminaries}

\textbf{Latent Diffusion Model with Control}. We use latent diffusion model (LDM)~\cite{rombach2022LDM} as the renderer module. LDM consists of two components: a variational autoencoder (VAE), which compresses input image to latent space with an encoder $z = E(I)$ and reconstructs latent features to image space with a decoder $I = D(z)$, and a 2D U-Net, which is trained by predicting the noise added to latent features at timestep $t \in (1, 2, \dots, T)$. Training loss for LDM is:
\begin{equation}
    \mathcal{L}_{\mathrm{LDM}} = \mathbb{E}_{\epsilon_t\in\mathcal{N}(0,1), t\in \mathcal{U}[0,T], c} \left[ || \epsilon_t - \epsilon_{\theta}(z_t ; t, c) ||^2 \right]
\end{equation}
where $z_t$ is the noisy latent at timestep $t$, $\epsilon_\theta$ is the noise prediction network to be trained, $c$ is the control for conditional generation. During inference, LDM generates images by iteratively removing U-Net-predicted noise from randomly sampled Gaussian noise for $T$ steps. For fair comparisons with existing works~\cite{gao2024magicdrive}, we adopt the pretrained Stable Diffusion v1.5 as our base LDM.

Aside from using text prompt guidance in the original LDM, Bench2Drive-R also incorporates pixel-space guidance using ControlNet~\cite{zhang2023controlnet}. ControlNet creates a trainable copy of the U-Net encoder. The outputs from each layer of the ControlNet are added to the outputs of the corresponding layer in the original U-Net encoders. ControlNet and U-Net are connected via zero-conv module to prevent random noise at the early stage of training.

\noindent\textbf{nuPlan Simulator and Benchmark}. nuPlan~\cite{karnchanachari2024nuplan} is a widely used reactive closed-loop planning benchmark based on large-scale real world data. nuPlan divides a long real-world driving journey into smaller, manageable driving scenarios. Each scenario has high-level navigation information such as goal points and route plans. It also contains sensor data collected along expert trajectories.


\subsection{Overall Framework} \label{General_Framework}

In Bench2Drive-R, we base our behavioral controller on nuPlan simulator~\cite{karnchanachari2024nuplan}, which keeps track of all the structural information of a driving scenario. At a given time $t$, the simulator is able to provide the following information:
\begin{enumerate}
    \item \textbf{3D bounding boxes and semantic labels}: $\mathbf{B_t} = \{(b_i, c_i)\}_{i=1}^{N_b}$, where $b_i = {(x_j, y_j, z_j)}_{j = 1} ^ 8 \in \mathbb{R}^{8\times 3}$ is the bounding boxes for both dynamic and static objects (cars, pedestrians, obstacles, etc.) within a specific range; $c_i\in \mathcal{C}_{box}$ is the semantic label. 
    \item \textbf{Vectorized map elements}: $\mathbf{M_t} = \{(v_i, c_i)\}_{i=1}^{N_m}$, where $v_i = {(x_j, y_j)}_{j=1}^{N_v}$ represents vertices for polygon map elements (roadblocks, cross-walk regions, etc.) and interior points for linestring map elements (lane dividers, stop lines, etc.); $c_i \in \mathcal{C}_{map}$ represents the map class. 
    \item \textbf{Ego states}: $\mathbf{E_t} \in \mathbb{R}^{N_e}$, including ego velocity, acceleration, steering angle, ego-to-global matrix etc.
    \item \textbf{Camera parameters}: $\mathbf{K} = \{\mathbf{K}_i\in \mathbb{R}^{4\times 4}\}_{i=1}^{N_{cam}}$, where $\mathbf{K}_i$ is the camera transformation matrix composed of intrinsic and extrinsic matrices that transforms points from Lidar coordinate system to image coordinate system.
    
    \item  \textbf{Original recorded sensor images with global ego coordinates}: $\{(\mathrm{coord}_{i}, \mathbf{I}_i)\}_{i=1}^{N_f}$, where $N_f$ is the number of total frames in current scenario; $\mathrm{coord}_{i}$ is the position of ego vehicle under global coordinate system; $I^{ref}_i \in \mathbb{R}^{N_{cam}\times C \times {H} \times{W}}$ represents sensor images collected by $N_{cam}$ cameras. Note that they are  recordings of human driving and could not be directly used during simulation since the evaluated E2E-AD methods could behave differently from experts.
\end{enumerate}

During simulation as shown in Fig.~\ref{fig:structure}, at each step, 
(1)The evaluated \textbf{E2E-AD agent  yields a planned trajectory} $\mathbf{Tr}_t = \{(x_i, y_i)\}_{i=1}^{T_f}$ based on current images $\mathbf{I_t}$ and ego states $\mathbf{E}_t$, where $T_f$ is the length of prediction. (2) Then, the \textbf{behavioral controller executes the predicted trajectory $\mathbf{Tr}_t$ and generates behaviors} of other driving agents to update bounding boxes, map elements, and ego status. In this work, we adopt nuPlan's rule-based IDM policy~\cite{Treiber_2000IDM} while it could also be learning-based models~\cite{li2024think2drive} for traffic simulation~\cite{karnchanachari2024nuplan}. (3) Finally, the proposed \textbf{generative renderer generates new surrounding images} $\mathbf{I}_{t+1}$ based on aforementioned information in the scene. The three steps are iteratively executed during the closed-loop simulation.

\subsection{Generative Renderer}
The key innovation of Bench2Drive-R is the generative renderer, which is composed of Latent Diffusion Model (we adopt pretrained Stable Diffusion v1.5) and  ControlNet. As shown in Fig.~\ref{fig:renderer}, we unify multiple simulation oriented designs into a ControlNet encoder to achieve controllable, consistent, and interactive surrounding image generation in AD scenarios based on given conditions from behavioral controller and database. We demonstrate the details in the following sections.

\begin{figure*}[!tb]
    \centering
\includegraphics[width=1.0\linewidth]{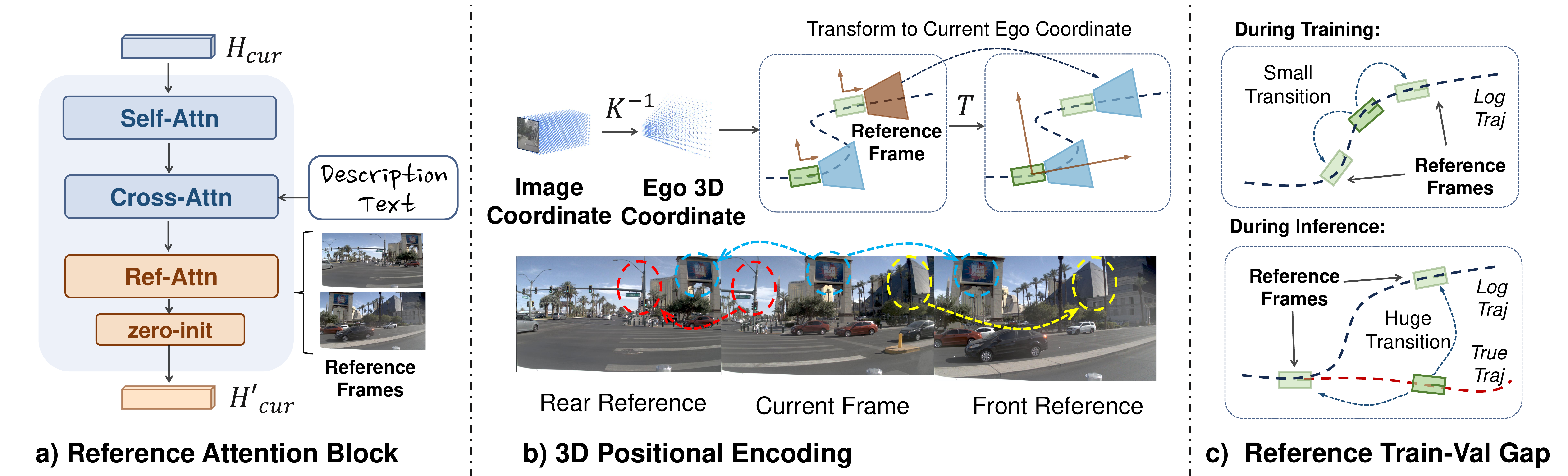}
\caption{\textbf{Designs for Retrieval based Scene-Level Control.} \textbf{a)} Reference images are utilized in ControlNet with an additional cross attention module (Ref-Attn).  \textbf{b)} Pixel-level 3D position encodings are calculated and fed into cross attention to provide spatial relations between reference and current images.   \textbf{c)} E2E-AD agents might not follow logged trajectory during inference, leading to train-val gap. }
    \label{fig:retrieval}
\end{figure*}

\subsubsection{Temporal Consistency Module \& Mitigation of Autoregressive Distribution Shift} \label{TCM}



Different from video diffusion models~\cite{blattmann2023SVD, blattmann2023AlignLatent, ho2022videodiffusionmodels} which improve temporal consistency by introducing attention along temporal axis of the noise, Bench2Drive-R, as an autoregressive interactive generation method, improves temporal consistency by encoding previously generated images $\mathbf{I}_{t-1}$ with ControlNet~\cite{zhang2023controlnet}. Specifically, We encode $\mathbf{I}_{t-1}$ into latent space with the same VAE encoder as Stable Diffusion's and send it into the ControlNet encoder. The output hidden features from each layer of ControlNet are directly added to the corresponding layers of the U-Net encoder, as in Fig.~\ref{fig:renderer}.
 

However, utilizing the previous image introduces a train-val gap issue. During training, the previous images are always ground-truth while during inference, previous images are from generation which have gaps with real world ones, even slightly. As a result, due to the recurrent nature of autoregressive generation, the error accumulates and could finally collapse the generation, as shown in Fig.\ref{fig:prev_mech} (Left). It is called teacher-forcing~\cite{lamb2016professor} or distribution shift~\cite{ross2010efficient} issue.

Since the deterioration stems from over-reliance on previous images~\cite{valevski2024gamengen},  we propose \textbf{noise modulation with Gaussian blur} to address the issue. Specifically, during training, previous-frame images are firstly encoded into latent space to get conditional previous latent $z_{\mathrm{prev}}$. Then a random level of Gaussian noise is added to $z_{\mathrm{prev}}$. The noise level is also input into the ControlNet encoder, which can be formulated as:
\begin{equation}
    c_{\text{prev}} = \mathcal{E}(\sqrt{\bar{\alpha}_n}z_{\mathrm{prev}} + \sqrt{1-\bar{\alpha}_n}\epsilon \; ;\; n)
\end{equation}
where $\mathcal{E}$ represents the ControlNet encoder; $\epsilon \in \mathcal{N}(0,1)$ is the randomly sampled Gaussian noise and $n\in \mathcal{U}[0, N]$. The noise-adding policy is similar to the training strategy of diffusion models~\cite{ho2020ddpm, song2022ddim}, where the noise level $n$ here is analogous to the timestep $t$ in diffusion models. To further avoid accumulation of high-frequency artifacts, we apply Gaussian blurring to previous-frame images. As shown in Fig.~\ref{fig:prev_mech} and later experiments, the proposed techniques could effectively alleviate the deterioration issue.


\subsubsection{Projected Object-Level Control} \label{POC}
This module follows existing generative models for AD~\cite{wang2023drivedreamer,wen2023panacea} and \textit{we do not claim it as our contributions}. During simulation, the object-level information $\mathbf{B}_t$ and $\mathbf{M}_t$ is available from behavioral controller. Thus, to adopt them as control information, we project 3D bounding boxes $\mathbf{B}_t$ and vectorized map elements $\mathbf{M}_t$ Lidar coordinate system to 2D perspective view using the provided camera parameters $\mathbf{P}$~\cite{gao2024magicdrive,wang2023drivedreamer,wen2023panacea}. Then we plot the projected discrete coordinates to form a set of binary masks of the same size with input images $\mathbf{B}_t^{\mathrm{mask}} \in \mathbb{R}^{|\mathcal{C}_\mathrm{box}| \times H \times W}$ and $\mathbf{M}_t^{\mathrm{mask}} \in \mathbb{R}^{|\mathcal{C}_{\mathrm{map}}| \times H \times W}$. We incorporate object semantic information by assigning each class its own dedicated channel. The two kinds of object level controls are concatenated and encoded into the latent space with a simple convolutional network. Then, these encoded control signals are injected into the denoising process using the same ControlNet encoder described in Section \ref{TCM}. Our projected object-level control can be formulated as:
\begin{equation}
    c_{\text{proj}} = \mathcal{E}(\mathrm{Conv}(\mathrm{Cat}(\mathbf{B}_t^{\mathrm{mask}}, \mathbf{M}_t^{\mathrm{mask}})))
\end{equation}

\subsubsection{Retrieval based Scene-Level Control} \label{RSC}

Generative models are susceptible to creating fictitious artifacts~\cite{oasis2024}. Previous studies on generative models for autonomous driving have primarily focused on generating a diverse range of driving scenes~\cite{gao2024magicdrive, wen2023panacea} to enhance the ground-truth dataset, effectively serving as a form of data augmentation. By contrast, \textbf{for the proposed simulation framework, fidelity is the most important factor for generative renderer since the tasks of enhancing diversity is assigned to the initial scenario selection and the behavioral controller}. In other words, we expect the diffusion models to faithfully follow controls and conditions, akin to a meticulous oil painter, rather than behaving like an artist who seeks to create diverse interpretations.

\begin{figure}[!tb]
    \centering
\includegraphics[width=1.0\linewidth]{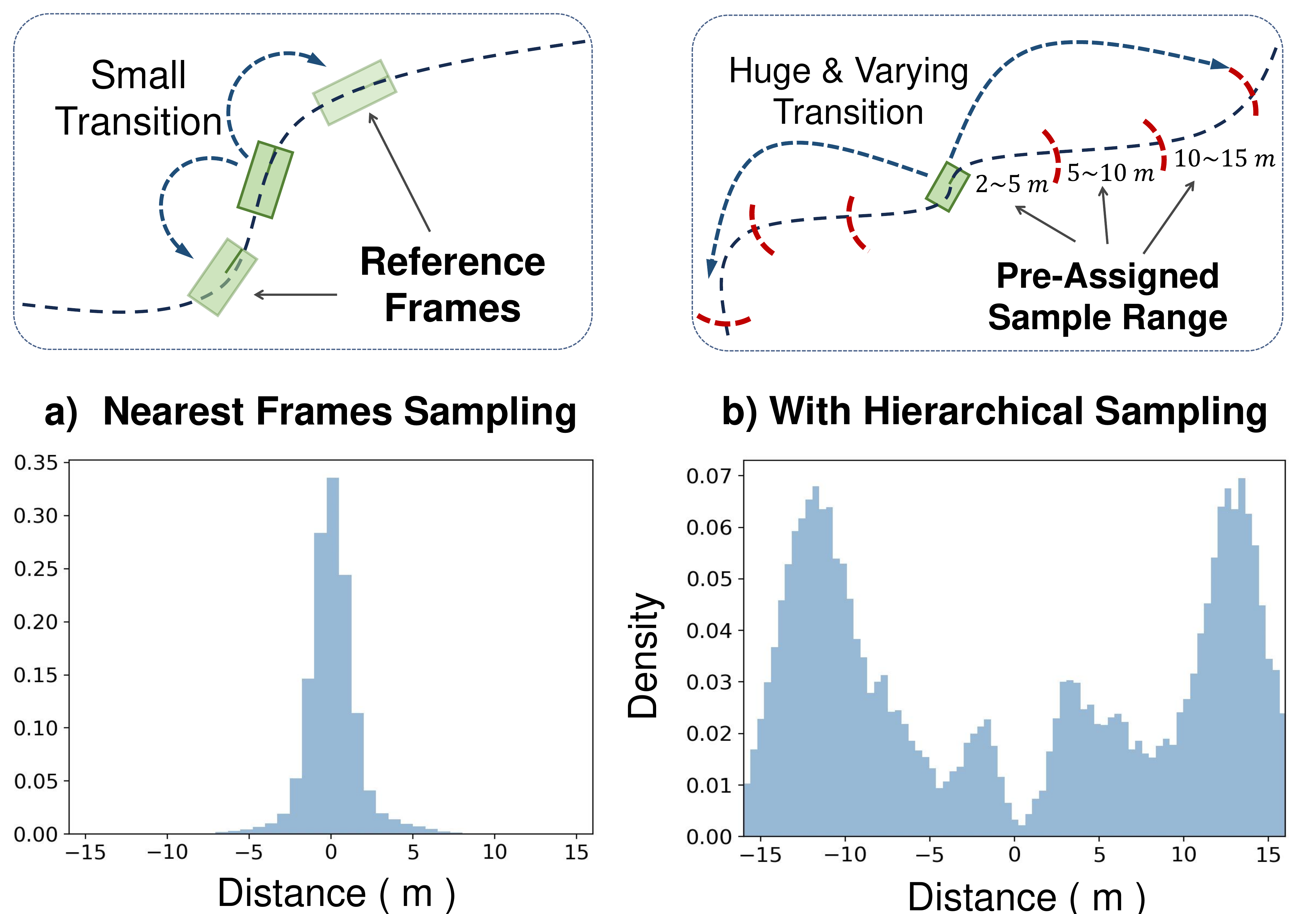}
    \caption{\textbf{Different Sampling Strategies for Reference Frames during Training.} \textbf{a}) Simply retrieving the closest one would make the model only be able to deal with very similar reference images. \textbf{b}) The proposed hierarchical sampling strategy forces the model adapts to reference images from a large scale of distance range.}
    \label{fig:histo}
    \vspace{-3mm}
\end{figure}

To achieve this, \textbf{one key observation is that the background of scene could be deterministically decided by referring to recordings $\{(\mathrm{coord}_{i}, \mathbf{I}_i)\}_{i=1}^{N_f}$ since the background is static}. Thus, for high fidelity, extra control conditions could be applied by retrieval to eliminate uncertainty. Specifically, \textbf{the two frames within recordings with closest distance} (one from ahead of the ego vehicle and one from behind) to current location of ego agent are retrieved:
\begin{equation}
\begin{aligned}
    \mathcal{P} = \{(\mathbf{coord}_i - &\mathbf{coord}_{\mathrm{ego}})\cdot \mathbf{v}_{\mathrm{ego}} \}_{i=1}^{N_f}\\
    \mathbf{I}_{\mathrm{ref}}^{\mathrm{front}} = \mathbf{I} \left[ \mathop{\mathrm{argmin}}\limits_{i; \; \mathcal{P}_i>0}\;\mathcal{P}_i\right] \; &; \; \mathbf{I}_{\mathrm{ref}}^{\mathrm{rear}} = \mathbf{I} \left[ \mathop{\mathrm{argmax}}\limits_{i; \; \mathcal{P}_i<0}\;\mathcal{P}_i\right]
\end{aligned}
\end{equation}
The two retrieved images are encoded by an image encoder (e.g., ResNet) and put into ControlNet as key and value in an additional cross-attention module so that the generation of current frames could find correspondence of background.

Further, since the spatial relations between current frame and the two reference frames could be explicitly calculated based on coordinate transformation, \textbf{we consider injecting pixel-wise spatial relations information into cross attention}. Specifically,  pixel-wise 3D position encodings are adopted similarly to~\cite{liu2022petrv1, liu2022petrv2}. A discrete meshgrid $\mathbf{P}$ of size $(H, W, D, 4)$ in camera frustum space is calculated for all images, where $D$ is the number of points sampled along the depth axis. Then, the meshgrid is transformed from camera frustum space to current ego coordinate system with camera transformation matrix $\mathbf{P}^{\mathrm{lidar}} = \mathbf{K}^{-1} \mathbf{P}$ for both current frame $\mathbf{P}_{\mathrm{ego}}$ and the two reference frames $\mathbf{P}_{\mathrm{ref}}$. Finally, $\mathbf{P}_{\mathrm{ego}}$  serves as the PE of query while $\mathbf{P}_{\mathrm{ref}}$ serves as PE of key within the cross attention of ControlNet (Fig.~\ref{fig:retrieval} (a)(b)):
\begin{equation}
        \mathrm{H}_{\text{cur}}^\prime = \text{Attn}(Q=\mathrm{H}_{\text{cur}}+\mathbf{P}_{\mathrm{ego}}, K=\mathrm{H}_{\text{ref}}+\mathbf{P}_{\mathrm{ref}}, V=\mathrm{H}_{\text{ref}})
\end{equation}
which enables thecurrent frame to find the correspondence in reference images so that it can follow the static background and generate the transformed pixels.




\begin{figure}[!tb]
    \centering
\includegraphics[width=1.0\linewidth]{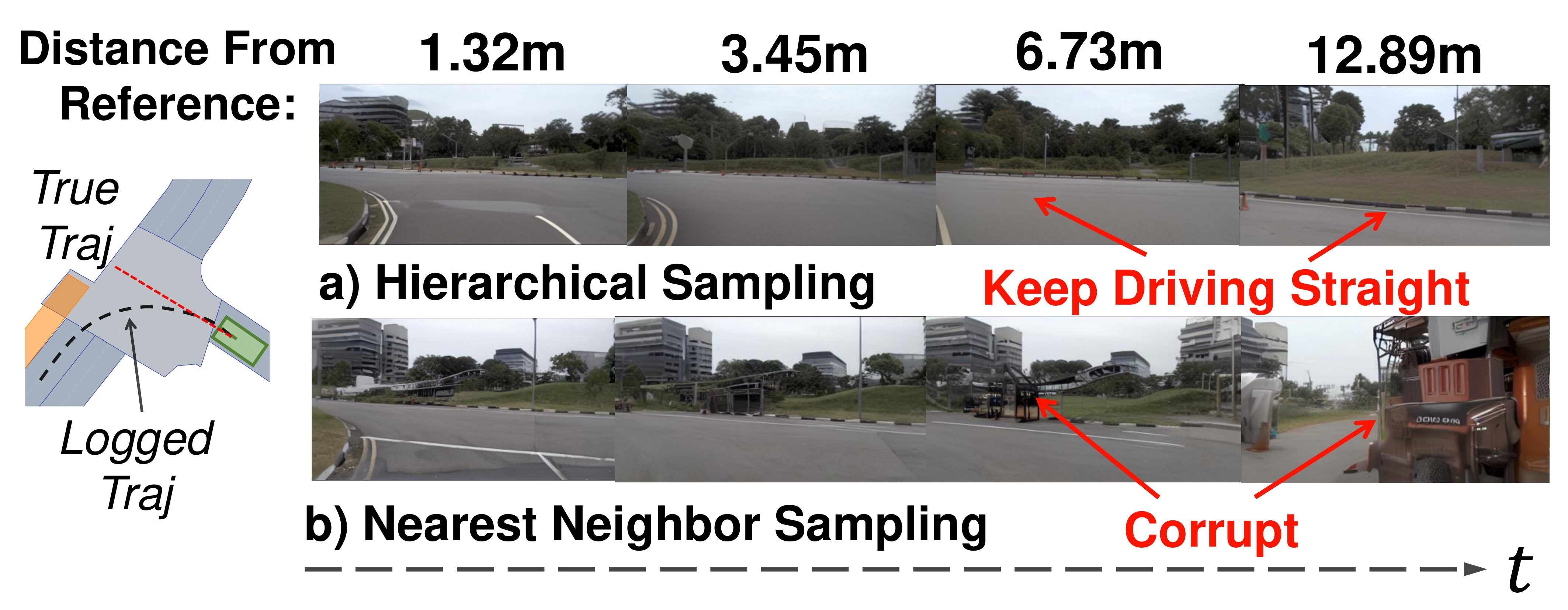}
    \caption{\textbf{Case Study on Influence of Different Training Sampling Strategies.} Hierarchical sampling strategy preserves generation quality even under large deviation during inference.}
\label{fig:hierarchical_sampling}
\end{figure}

\begin{figure}[!tb]
    \centering
\includegraphics[width=1.0\linewidth]{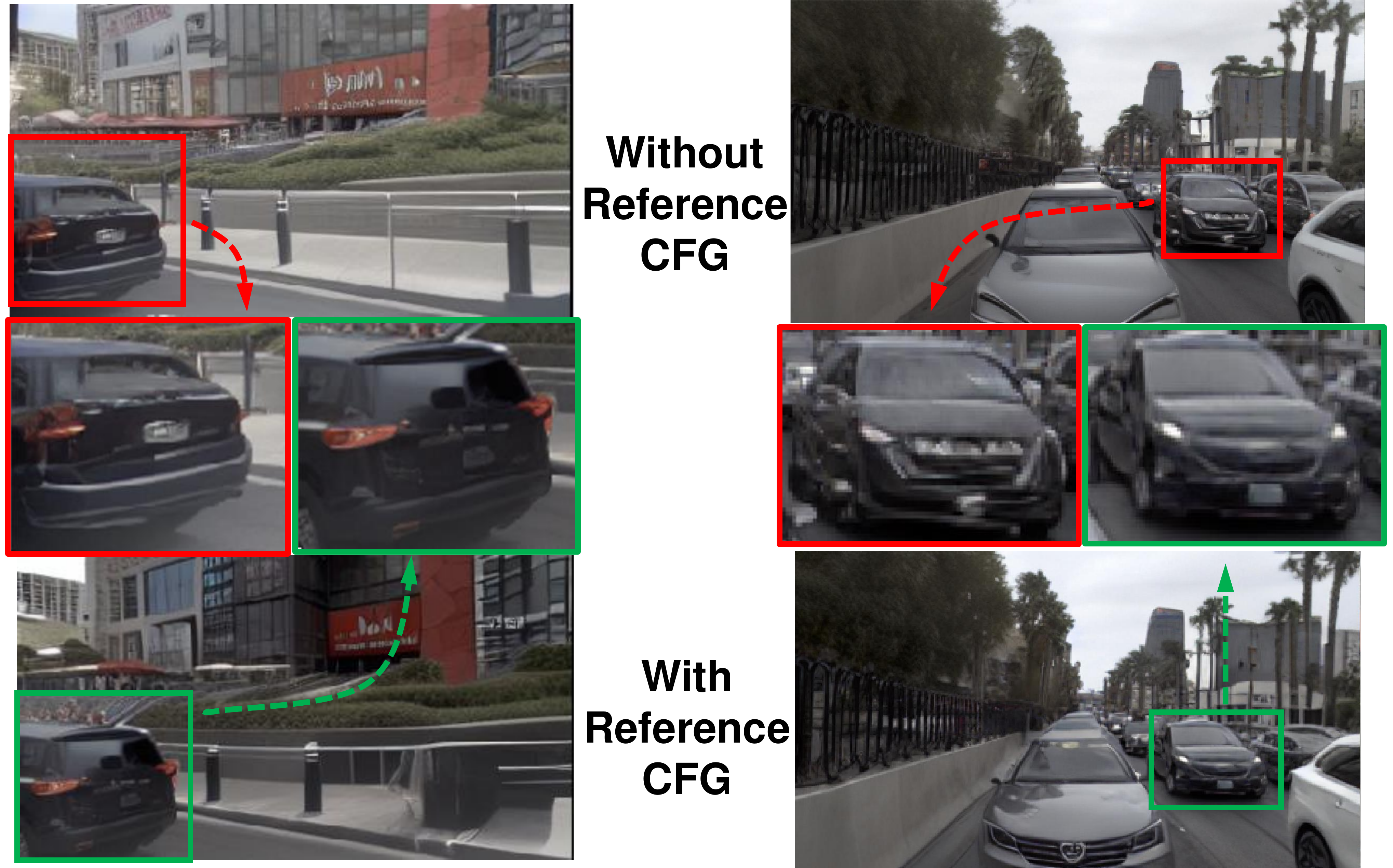}
    \caption{\textbf{Reference CFG} lessens reliance on reference frames and enhances foreground object authenticity.}
    \label{fig:ref_cfg_abla}
    \vspace{-3mm}
\end{figure}

By utilizing retrieval-based conditions, the street scene on both sides of the ego vehicle is deterministic and the generative renderer is only responsible for generating coherent images. However, this introduces another train-val gap challenge. \textbf{During training, if we simply retrieve the nearest images, they would always be the preceding and following frames within a small distance range while during inference, E2E-AD agents could behave differently from experts and thus the distance to reference images could be far,} shown in Fig.~\ref{fig:retrieval} (c). As a result, the training would cause the model to overly rely on references and collapse in the large deviation situations during inference.

To address the issue, we propose to \textbf{let the model see reference images in a wide range of distance during training}. Specifically, for each training sample, we employ a hierarchical sampling technique where reference images are selected out of one of the three distance intervals (2m-5m, 5m-10m, or 10m-15m) based on a pre-assigned probability to balance the sampling. In this work, we assign the probability to be (0.1, 0.3, 0.6), significantly increasing long-distance samples compared with simple nearest frames sampling, as in Fig.~\ref{fig:histo}. The proposed sampling makes the model adapt to highly divergent reference images, effectively narrowing the train-val gap and notably improving the generation quality and model's numerical stability. Fig.~\ref{fig:hierarchical_sampling} gives an example of long-range deviation and the generation results under different training strategies.

\begin{table*}[tb!]
    \centering
    \caption{Comparisons of FID, 3D object detection and BEV segmentation on nuScenes validation set. * means our replication.}
    \resizebox{0.9\linewidth}{!}{
    \begin{tabular}{l|c|ccc|cccc|cc}
    \hline \toprule
    \multirow{2}{*}{Method}   & \multirow{2}{*}{FID}  & \multicolumn{3}{c|}{ BEVFormer~\cite{li2022bevformer}} & \multicolumn{4}{c|}{ BEVFusion~\cite{liu2024bevfusion} (Camera Branch)}  & \multicolumn{2}{c}{StreamPETR~\cite{wang2023streampetr}}  \\ 
    &  & NDS$\uparrow$ & mAP$\uparrow$ & mAOE$\downarrow$ & NDS$\uparrow$ & mAP$\uparrow$ & mAOE$\downarrow$ & mIoU $\uparrow$ &NDS$\uparrow$ & mAP$\uparrow$ \\
    \midrule
    Oracle &- & 53.50 & 45.61 & 0.35  & 41.20 & 35.53 & 0.56 & 57.09 &57.10 &48.20 \\
    BEVControl~\cite{yang2023bevcontrol}& 24.85 & 28.68  & 19.64 & 0.78	& - & - & - & -	 & - & -	\\
    MagicDrive*~\cite{gao2024magicdrive} & 16.20 & 25.76  & 14.07	&0.79 &23.35 	&12.54 	& 0.77	&28.94 & 35.51 & 21.41\\
    Panacea~\cite{wen2023panacea} & 16.69 & -& -& -& -& -& -& -& 32.10 & - \\
    Panacea+~\cite{wen2024panaceaplus} & 15.50 & -& -& -& -& -& -& -& 34.60 & - \\
    Bench2Drive-R & \textbf{10.95}& \textbf{34.70}  & \textbf{20.11}	&\textbf{0.48}  & \textbf{25.75} & \textbf{13.53}  & \textbf{0.73} & \textbf{42.75} & \textbf{40.23} & \textbf{24.04}
    \\ 
    \bottomrule \hline
    \end{tabular}
    }
    \label{perception}
\end{table*}

\begin{table*}[tb!]
    \centering
    \caption{Performance of UniAD's Different Tasks in nuScenes. * means our replication.} 
    \resizebox{\textwidth}{!}{
    \begin{tabular}{l|cc|cccc|cc|c}
    \hline \toprule
    \multirow{2}{*}{Method}   & \multicolumn{2}{c|}{Detection}  & \multicolumn{4}{c|}{BEV Segmentation}  & \multicolumn{2}{c|}{Planning} & Occupancy  \\ 
    & NDS~\cite{nuscenes}$\uparrow$ & mAP$\uparrow$ & Lanes$\uparrow$ &Drivable$\uparrow$ & Divider$\uparrow$ & Crossing$\uparrow$ & avg.L2(m)$\downarrow$& avg.Col.$\downarrow$ & mIoU$\uparrow$ \\
    \midrule
    Oracle & 49.85 & 37.98 & 31.31  & 69.14  & 25.93  & 14.36 &  1.05 &  0.29  & 63.7 \\ 
    MagicDrive*~\cite{gao2024magicdrive} &	29.35 & 14.09&	23.73&	55.28&	18.83&	6.57&1.18 & 0.33	& 54.6 \\ 
    Bench2Drive-R&	\textbf{33.04} & \textbf{15.16}&	\textbf{25.5}&	\textbf{56.53}&	\textbf{21.27}&	\textbf{8.67}&	\textbf{1.15}&	\textbf{0.31}&	\textbf{55.5} 
    \\ \bottomrule \hline
    \end{tabular}
    }
    \label{tab:open-loop_planning}
\end{table*}

Further, we observe that the generation process might overly rely on reference images and thus leads to bad generalization ability. To alleviate the issue, we apply classifier-free guidance (CFG)~\cite{ho2022cfg} to reference image, dubbed reference CFG, where we randomly substitute reference images with empty images during training. Similar to standard CFG, at each denoising step during inference, we weighted sum the two predicted noise, one with reference images and one without. As shown in Fig.~\ref{fig:ref_cfg_abla}, reference-CFG alleviates model's reliance on reference image, resulting in more authentic foreground objects generation.

\section{Experiment}
\subsection{Experimental Setups}
\subsubsection{Dataset}

 We conduct detection evaluation and open-loop planning evaluation on \textbf{nuScenes dataset}~\cite{nuscenes} and closed-loop planning evaluation on \textbf{nuPlan dataset}~\cite{karnchanachari2024nuplan}. 
 

We use the nuScenes's official train-val-test split while for nuplan, we use the mini split (around 5 times larger than nuScenes) due to limited computational resource.

\subsubsection{Training and Inference}
We base the behavior controller in our framework on nuPlan~\cite{karnchanachari2024nuplan} simulators and our generative renderer on SDv1.5~\cite{Rombach_2022_CVPR}. Pretrained weights are used to initialize the U-Net layers.

At the training stage, we optimize our renderer for 50k steps with a total batch-size of 114 on nuScenes, and for 140k steps with a total batch-size of 64 on nuPlan. The learning rate is set to be $1e-4$ and cosine-annealing scheduler is employed with warm-up steps to be 3k steps. We assign the drop-out rate of retrieved reference images to be $0.2$, and Reference CFG's guidance scale is set to be $2$ at inference time.

During inference, following~\cite{gao2024magicdrive}, images are sampled using the UniPC~\cite{zhao2023unipc} scheduler for 20 steps. All sensor images are sampled at a spatial resolution of $400\times224$ and then upsampled to the original size with bicubic~\cite{bicubic} interpolation, namely $1600\times900$ for nuScenes and $2000 \times 1200$ for nuPlan.  We use the official UniAD pretrained weight on nuScenes and train an 8-views version of VAD on nuPlan dataset. Although the simulation frequency of nuPlan simulator is 10Hz, we follow the convention in the community and set the inference frequency of VAD to be 2Hz, and use the most recently predicted trajectory to propagate world state at the intermediate frames.

\begin{table*}[!tb]
\fontsize{8.5pt}{11pt}\selectfont
\setlength{\tabcolsep}{2.08mm}
\setlength{\aboverulesep}{0.4ex}
\setlength{\belowrulesep}{0.4ex}
\setlength{\abovecaptionskip}{1.5mm}
\centering
\begin{minipage}[t]{0.36\linewidth}
\centering
\setlength{\tabcolsep}{3pt}
\caption{Open-Loop Planning Ablation}
\resizebox{0.99\linewidth}{!}{
\begin{tabular}{c@{\hspace{0.2pt}}c|c|c|c}
\hline \toprule
\small{Temporal} & \small{Retrieval} & \multirow{2}{*}{FID} & Detection & Planning \\ 
\small{Consistency} & \small{Ref} & & NDS$\uparrow$ & avg.L2(m)$\downarrow$\\
\midrule
\XSolidBrush & \XSolidBrush & 21.06 &21.80  & 1.19  \\
\checkmark & \XSolidBrush & 14.04 & 25.75 &  1.17  \\
\checkmark & \checkmark & \textbf{10.95} &\textbf{33.04}&	\textbf{1.15}
\\ \bottomrule \hline
\end{tabular}}
\label{tab:oriented-abla}
\end{minipage}
\quad
\begin{minipage}[t]{0.30\linewidth}
\centering
\caption{Noise \& Gaussian Blur Ablation} 
\setlength{\tabcolsep}{3pt}
\resizebox{0.99\linewidth}{!}{
\begin{tabular}{c@{\hspace{-0.5pt}}c|cccc}
\hline \toprule
 Noise  & Gaussian & \multicolumn{4}{c}{FID}\\ 
 Modulation &  Blur & 0.5s & 1s & 1.5s & 2s \\
\midrule
\XSolidBrush & \XSolidBrush & \textbf{12.68} & 16.09 & 23.77&  31.79 \\ 
\checkmark & \XSolidBrush  & 14.87 & 17.64 & 18.72&	19.58\\ 
\checkmark & std=1 &	14.40 & \textbf{16.84} &	\textbf{18.01}&	\textbf{18.67} \\
\checkmark & std=2 &	15.55 & 18.14 &	19.11&	19.79
\\ \bottomrule \hline
\end{tabular}}
\label{tab:prev_abla}
\end{minipage}
\quad
\begin{minipage}[t]{0.28\linewidth}
\setlength{\tabcolsep}{2pt}
\caption{Closed-Loop Planning } 
\centering
\resizebox{0.99\linewidth}{!}{
\begin{tabular}{l|cc|c}
    \hline \toprule
     \multirow{2}{*}{Methods}  & \multicolumn{2}{c|}{BEVFormer} & \multirow{2}{*}{R-CLS}\\ 
      &  NDS$\uparrow$ & mAP$\uparrow$ &  \\
    \midrule
    Log-Replay & 0.05 & 0.03 & 27.24 \\ 
    No Ref \& No Prev & 23.31 & 12.21 & 28.56\\ 
    Bench2Drive-R & \textbf{28.23} & \textbf{17.23} &	\textbf{30.49} \\ \bottomrule \hline
    \end{tabular}}
    \label{tab:closeloop}
\end{minipage}
\end{table*}

\begin{figure*}[!tb]
    \centering
\includegraphics[width=1.0\linewidth]{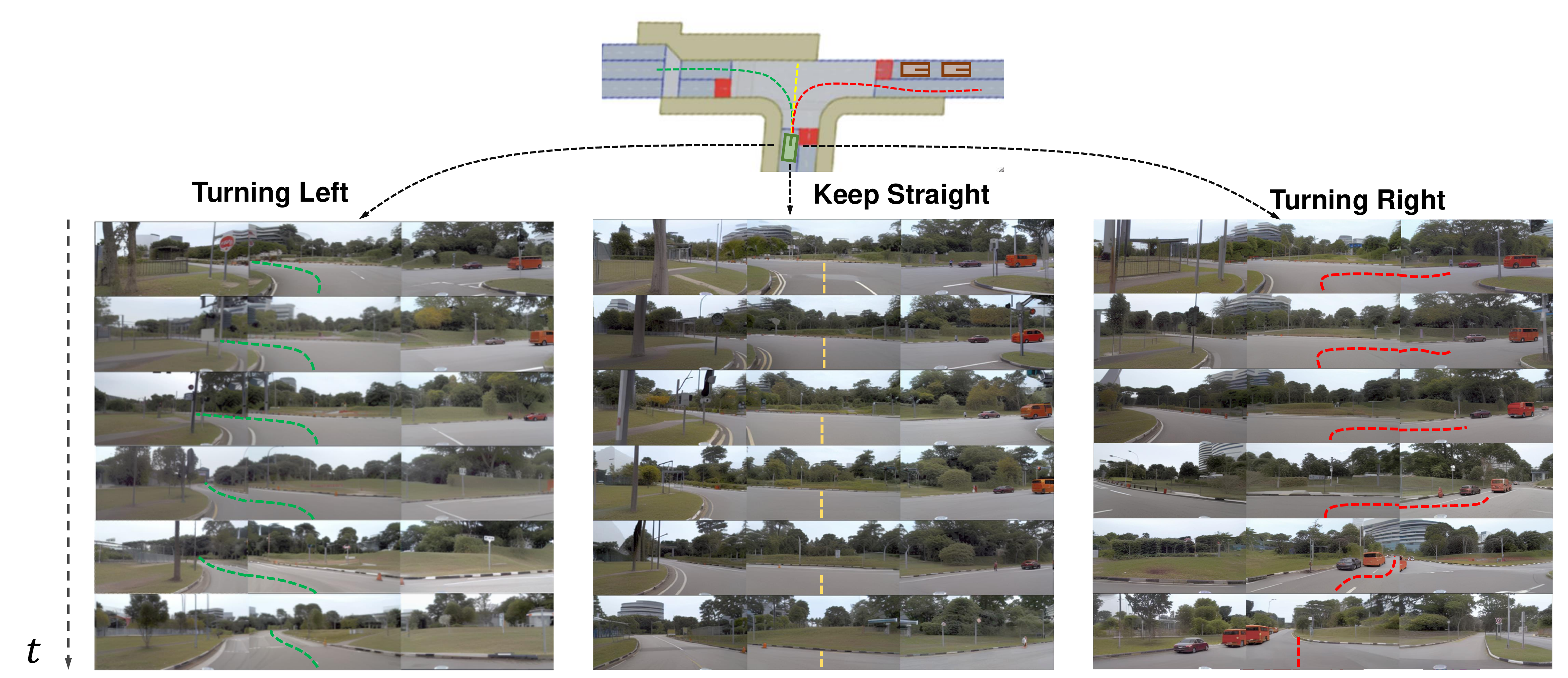}
    \caption{\textbf{Closed-Loop Interactive Simulation in nuPlan.} Generated image sequences under three different E2E-AD agent behaviors.}
    \label{fig:predictive}
\end{figure*}

\subsubsection{Metrics}
Following existing works, we evaluate the \textbf{generation quality} with Frechet Inception Distance (FID)~\cite{heusel2017gans}, which measures the distance between the distributions of real and generated images, reflecting image synthesis quality. The \textbf{layout conformity} of Bench2Drive-R generative renderer is evaluated through performing object detection and BEV segmentation on the generated images. Widely used baselines BEVFormer~\cite{li2022bevformer} and BEVFusion~\cite{liu2024bevfusion} (camera branch) are selected. To evaluate the \textbf{temporal consistency} of the generated image sequence, we employ state-of-the-art streaming perception models StreamPETR~\cite{wang2023streampetr}. StreamPETR features the reuse of agent queries from previous frames and thus improved perception scores indicate better temporal consistency. To evaluate the influence on \textbf{planning}, we employ UniAD~\cite{hu2023planning} and  VAD~\cite{jiang2023vad} for open-loop and closed-loop evaluation respectively. 

For closed-loop evaluation, we employ \textbf{CLS} (Closed-Loop Score) defined by the official nuPlan challenge. CLS is a scenario-based metric, which comprehensively combine multiple aspects of driving performance assessments including  drivable area compliance, collision time, progress along the driving direction, comfort, etc.

\subsection{Main Results}

\subsubsection{Quantitative Analysis}

\noindent\textbf{Generation Quality and Controllability}. We evaluate the generation capability of Bench2Drive-R's generative renderer with nuScenes validation dataset. As shown in Tab.~\ref{perception}, Bench2Drive-R outperforms baselines BEVControl~\cite{yang2023bevcontrol}, MagicDrive~\cite{gao2024magicdrive}, Panacea~\cite{wen2023panacea} and Panacea+~\cite{wen2024panaceaplus} in generation quality, yielding notably lower FID score. For controllability, better perception and segmentation scores are achieved on Bench2Drive-R-generated images, indicating better generation precision for objects and map elements.  

\noindent\textbf{Temporal Consistency.} Bench2Drive-R can yield consistent sensor image sequence over a long horizon, which is crucial for perception models with high temporal reliance. As shown in Tab.~\ref{perception}, perception scores with StreamPETR are notably better than the baseline method. This demonstrates the effectiveness of our method in improving temporal consistency under autoregressive generation setting.

\noindent\textbf{Open-Loop Evaluation.} As in Tab.~\ref{tab:open-loop_planning}, UniAD performs better on Bench2Drive-R generated image sequences than baselines under nuScenes open-loop evaluation protocol.

\noindent\textbf{Closed-Loop Planning.} We integrate the Bench2Drive-R framework into nuPlan for \textbf{closed-loop reactive simulation}. We adopt the Val14 evaluation split~\cite{dauner2023partingmisconceptionslearningbasedvehicle}. However, since only 10\% scenes in nuPlan have sensor data, we filter 10 full clips from each of the 14 scenarios and report R-CLS score and perception score. We use two simple image acquisition methods to serve as baselines, log replaying (collecting images with corresponding timestep from recorded ego trajectories) and static frame generation with no previous prior or reference frames. 

As is shown in Tab.~\ref{tab:closeloop}, Bench2Drive-R's perception scores are notably higher than baseline methods, indicating sensor images' great adherence to bbox-level simulation environment and the efficacy of our methods. However, R-CLS of VAD exhibits only a marginal improvement. This is because VAD, which adopts an imitation learning paradigm, is not capable of coping with long horizon closed-loop simulation, which is aligned with previous findings~\cite{cheng2023plantf}. Driving scores of VAD tend to drop to zero at the early stage of simulations due to its limited capability. We provide more case studies for VAD planning ability in Section~\ref{VAD}.

\subsubsection{Qualitative Analysis}

\noindent\textbf{Closed-Loop Interactive Simulation}. As in Fig.~\ref{fig:predictive}, 
 Bench2Drive-R is able to generate high-fidelity images under different behaviors of E2E-AD agents.

\begin{figure}[!tb]
    \centering
\includegraphics[width=1.0\linewidth]{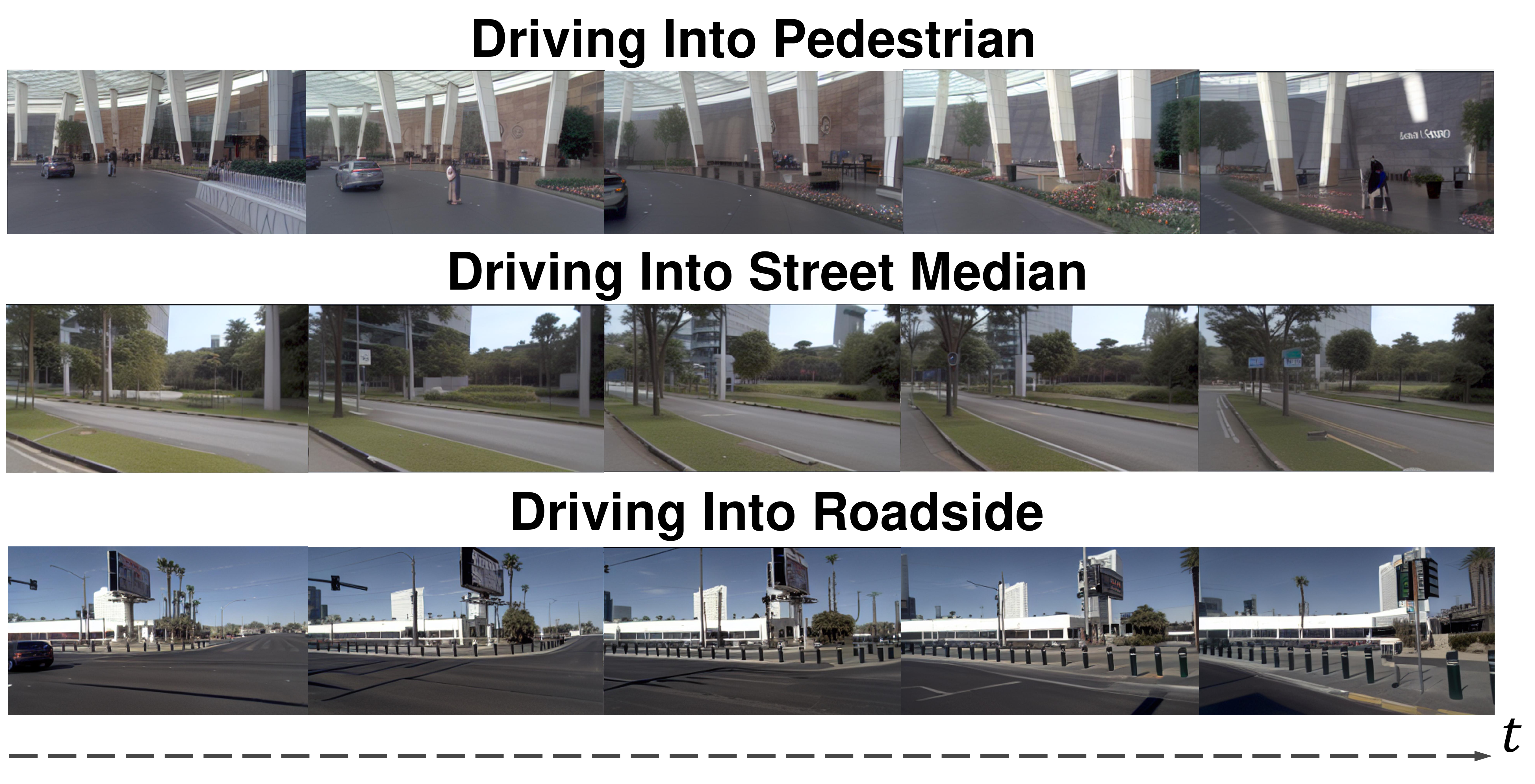}
\caption{\textbf{Generalizability.} Out-of-distribution generation results under scenarios absent in the training dataset. }
    \label{fig:ood}
\end{figure}

\noindent\textbf{Generalizability}. As in Fig.~\ref{fig:ood}, Bench2Drive-R can generate authentic sensor images even under scenarios absent in training dataset, demonstrating the rich real-world prior knowledge in the pretrained diffusion model.


\subsection{Ablative Study}
\textbf{Designs of Generative Renderer.} In Tab.~\ref{tab:oriented-abla}, we ablate the two proposed simulation oriented designs and results show that task performance improves along with the modules we add, demonstrating their effectiveness.

\noindent\textbf{Noise Modulation with Gaussian Blurring}
As shown in Tab.~\ref{tab:prev_abla}, directly encoding previous images leads to fast deterioration while with noise modulation and a proper level of Gaussian noise added, generalization quality remains stable during autoregressive generation process.

\begin{figure}[!tb]
    \centering
\includegraphics[width=1.0\linewidth]{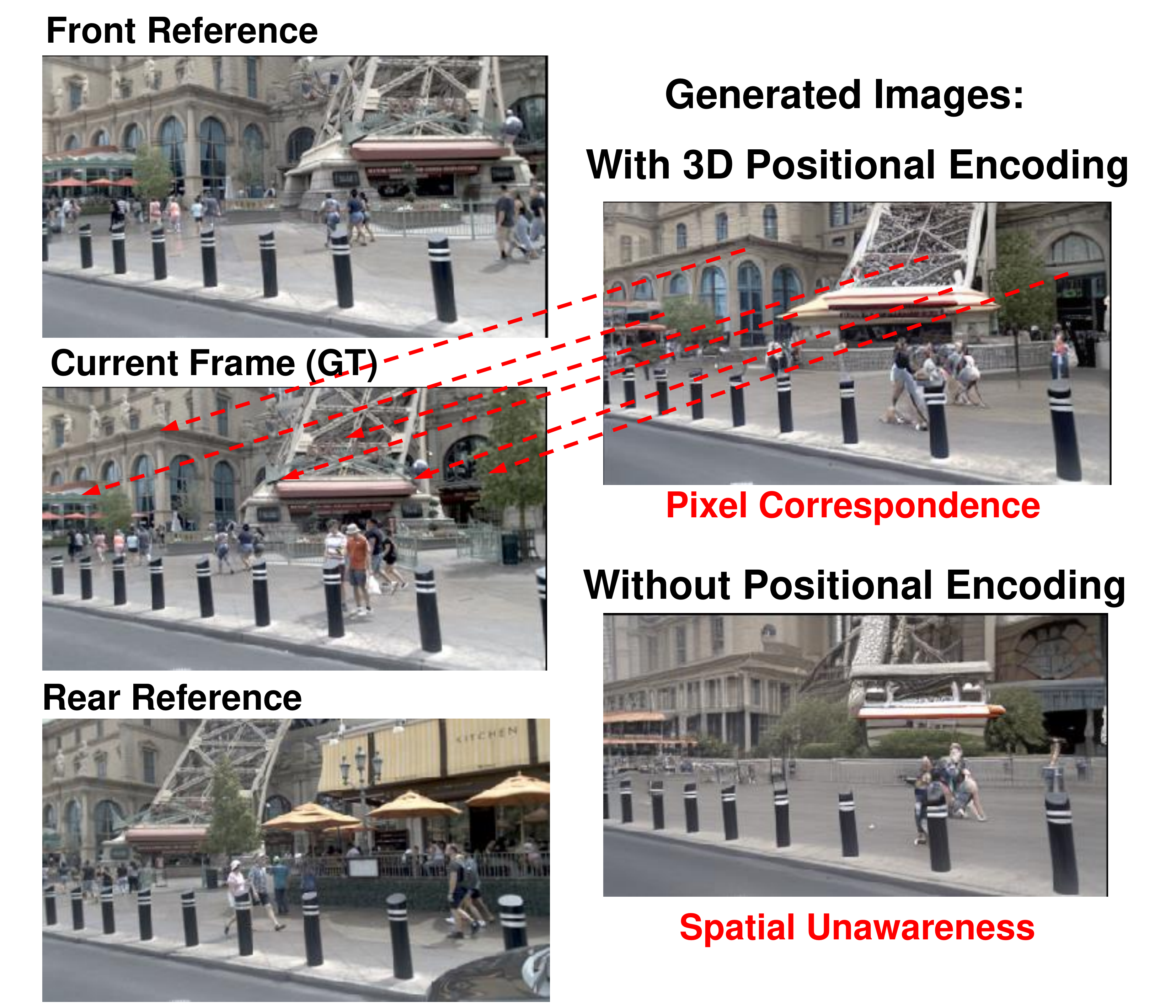}
    \caption{\textbf{Effect for 3D Positional Encoding.} 3D PE provides explicit spatial transition information, which effectively avoids spacial unawareness. }
    \label{fig:ref_pe_abla}
\end{figure}

\noindent\textbf{3D Positional Encoding}. As shown in Fig.~\ref{fig:ref_pe_abla}, with the explicit spatial transformation information provided by 3D PE, generated images exhibit pixel-level correspondence.


\subsection{Analysis on VAD Planning Performance}\label{VAD}
\begin{figure*}[!tb]
    \centering
\includegraphics[width=0.7\linewidth]{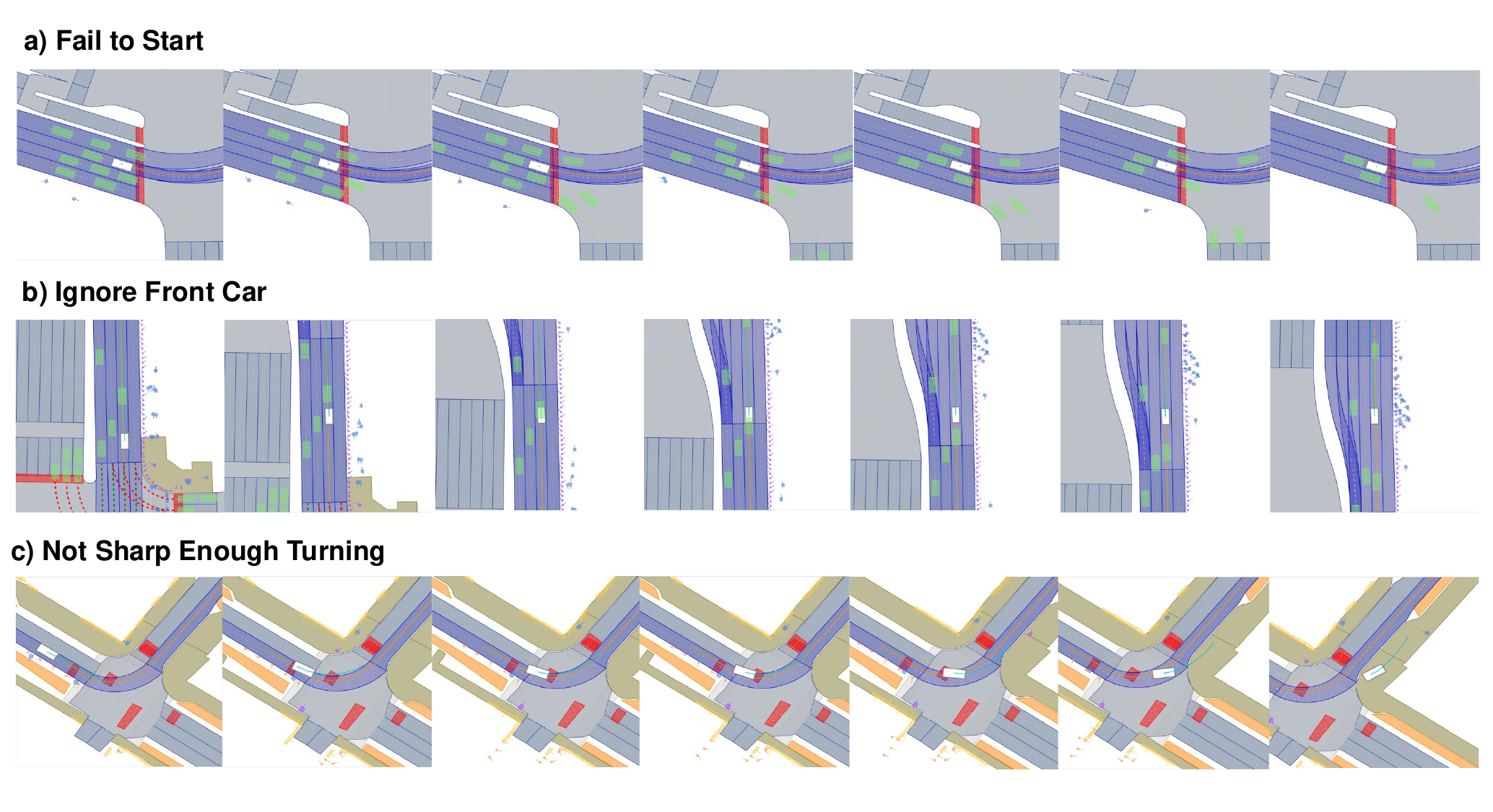}
    \caption{\textbf{Typical Failure Cases of VAD} We select three typical failure cases of VAD: failing to start, accelerating when there are cars in the front and failing to take turns. The white box is the ego car; green boxes are other driving cars; the green line is the planned ego trajectory and the orange line is logged expert trajectory.}
    \label{fig:failure}
\end{figure*}

We provide some typical failure cases of VAD in Fig.~\ref{fig:failure}. As suggested by Fig.~\ref{fig:failure} (a) and (b), VAD is hard to start from static states, and is unable to slow down even when there are slow cars in the front. This proves the findings~\cite{cheng2023plantf, zhai2023rethinking} that imitation based driving models are likely to take shortcuts from current kinetic states during training, developing an overly reliance on ego states while ignoring other information during inference.

In Fig~\ref{fig:failure} (c), VAD's planning results are very close to ground-truth trajectory at the early stage of a left turn. However, the ego car gradually deviates from the original route because of accumulated errors and the model can't adapt to the changes and hit the roadblock. This result intermediately reflects that the now commonly-used open-loop protocols is not capable of evaluating model's true driving ability.

\section{More Qualitative Results}

\subsection{Generation Quality}
\begin{figure*}[p]
    \centering
\includegraphics[width=0.95\linewidth]{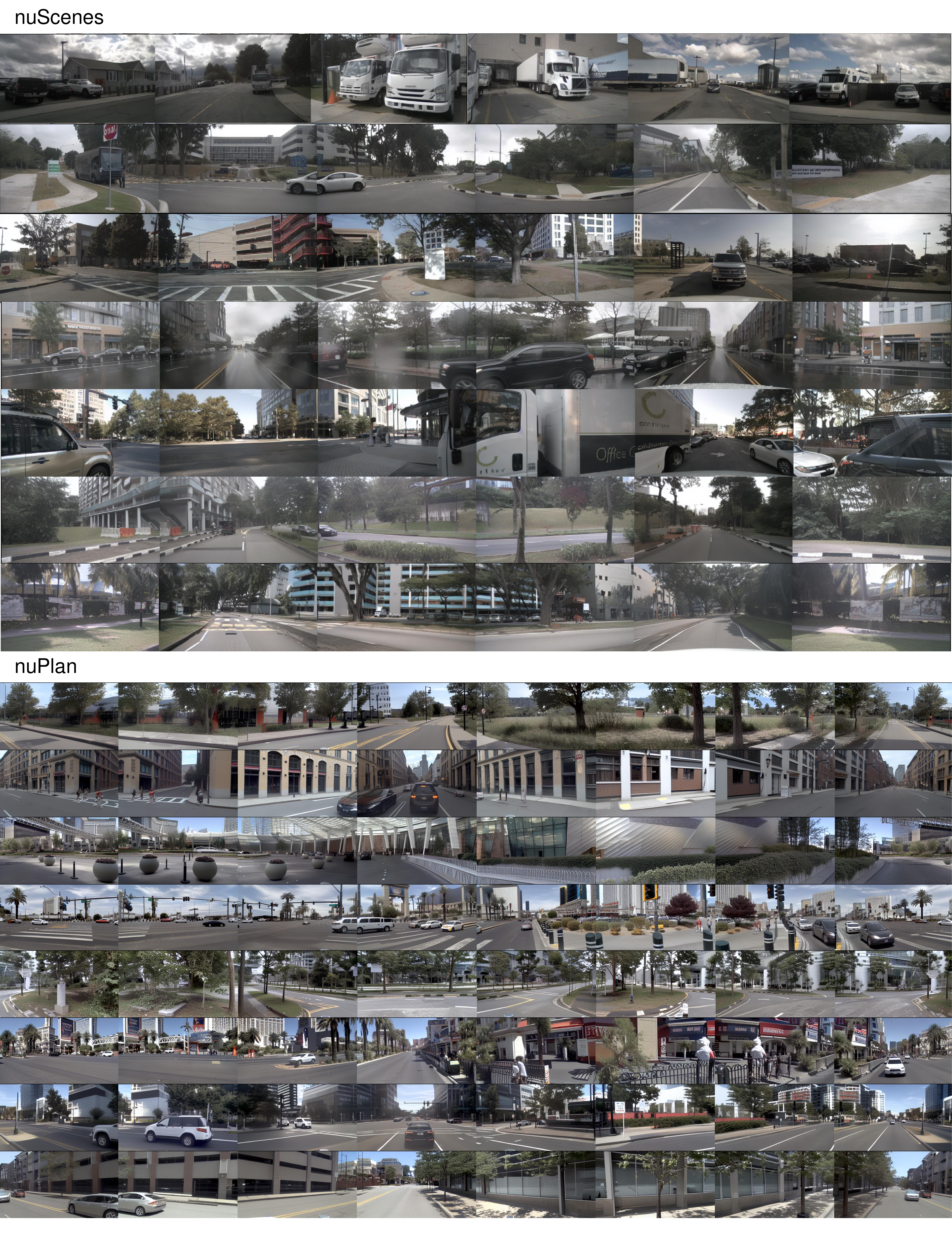}
    \caption{\textbf{Generated Images In NuScenes and NuPlan.} Bench2Drive-R is capable of generating diverse driving scenarios with high fidelity.}
    \label{fig:quality}
\end{figure*}

As shown in Fig.~\ref{fig:quality}, Bench2Drive-R is capable of generating high-fidelity panoramic images under diverse driving scenarios.

\subsection{Controllability \& Spacial Consistency}

\begin{figure*}[!tb]
    \centering
\includegraphics[width=0.8\linewidth]{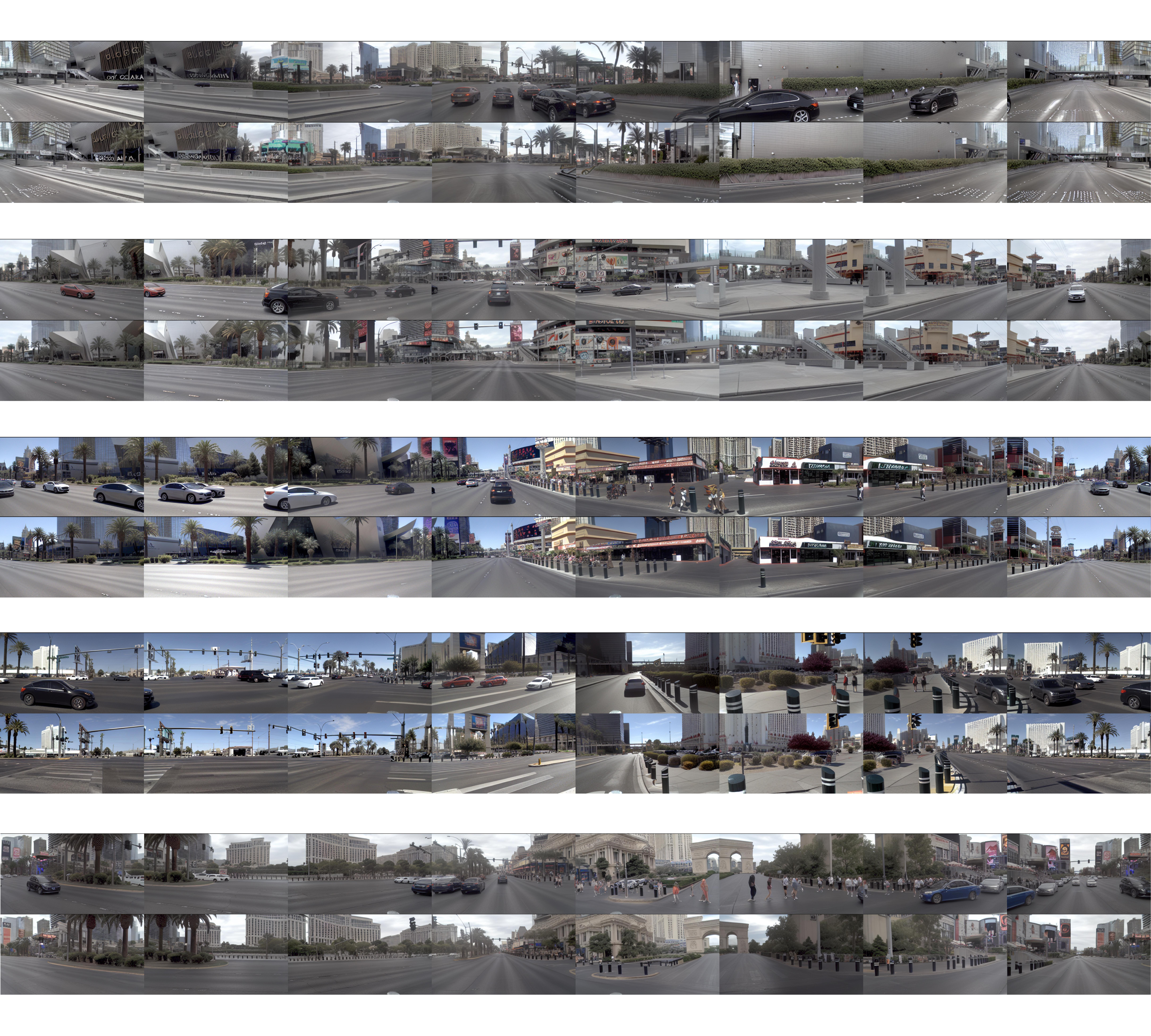}
    \caption{\textbf{Controllability and Spatial Consistency} For each set of images, the upper row is generated with object bounding boxes and the lower row without. Bench2Drive-R abides strictly by the control signals while maintaining high background consistency.}
    \label{fig:nobox}
\end{figure*}

We demonstrate the controllability of Bench2Drive-R by removing all object bounding boxes in a driving scenarios, as shown in Fig.~\ref{fig:nobox}. For each scenario, we generate two sets of images, one with bounding boxes and one without. Despite the radical changes of object control signals, Bench2Drive-R is able to achieve high spacial consistency at the background level and removes all foreground objects in the scenario (including driving cars and pedestrians), demonstrating the efficacy of our designs for the generative renderer. 

\subsection{More Interactive Simulation Visualization}
\begin{figure*}[p]
    \centering
\includegraphics[width=0.94\linewidth]{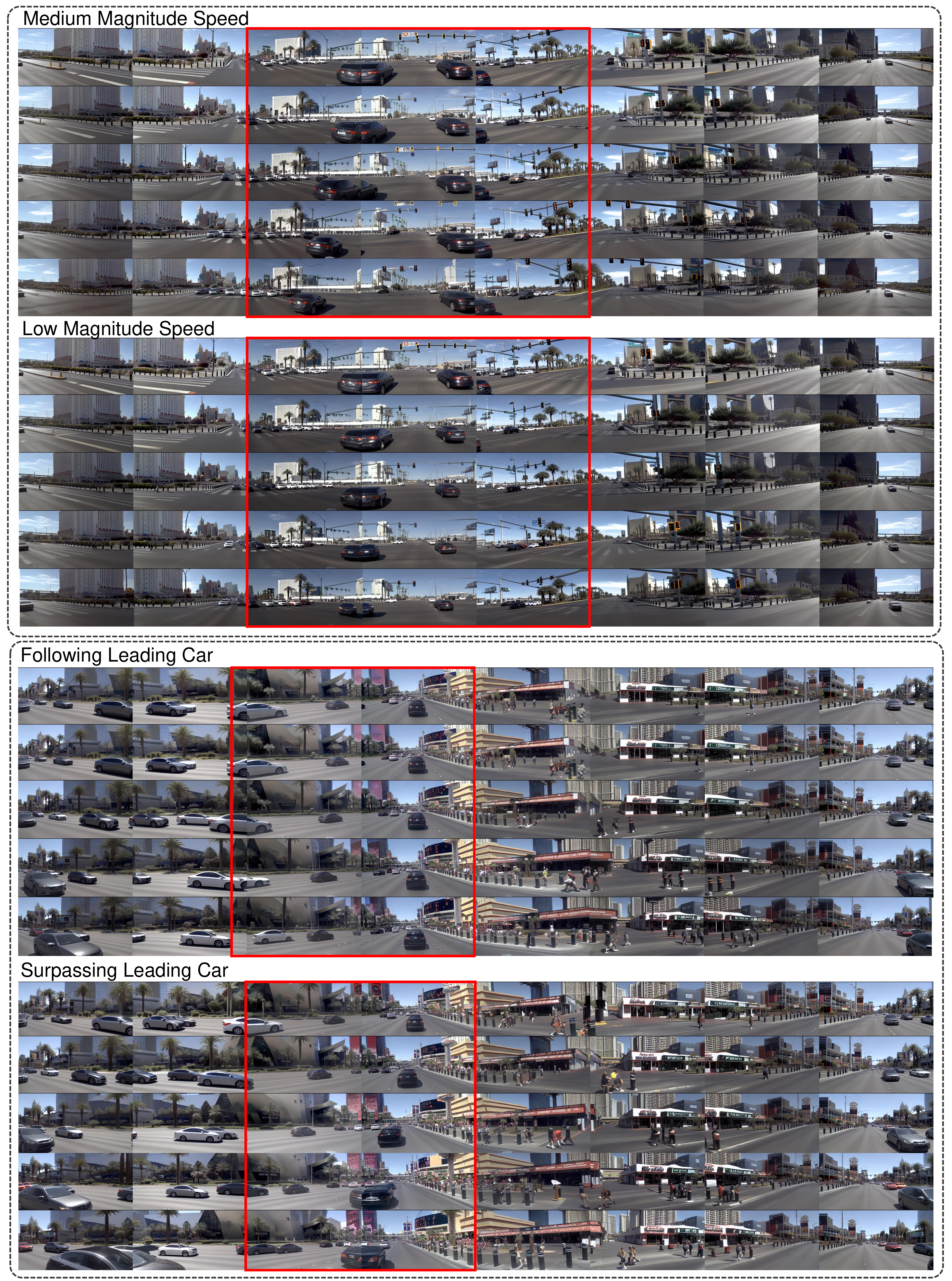}
    \caption{\textbf{Interactive Simulation Result}. Views with most conspicuous differences are highlighted with red boxes.}
    \label{fig:inter1}
\end{figure*}

\begin{figure*}[p]
    \centering
\includegraphics[width=0.94\linewidth]{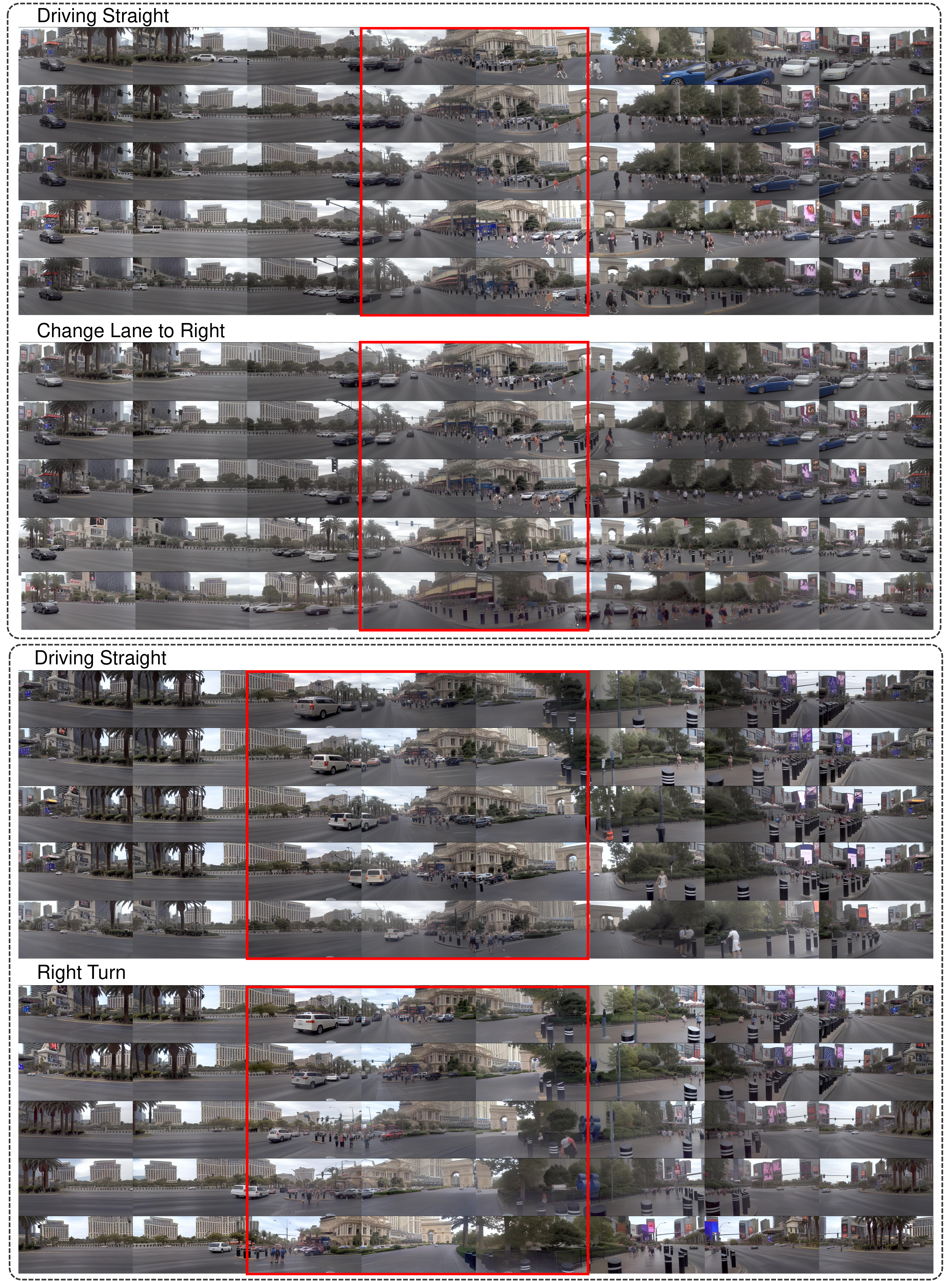}
    \caption{\textbf{Interactive Simulation Result}. Views with most conspicuous differences are highlighted with red boxes.}
    \label{fig:inter2}
\end{figure*}

\begin{figure*}[p]
    \centering
\includegraphics[width=0.94\linewidth]{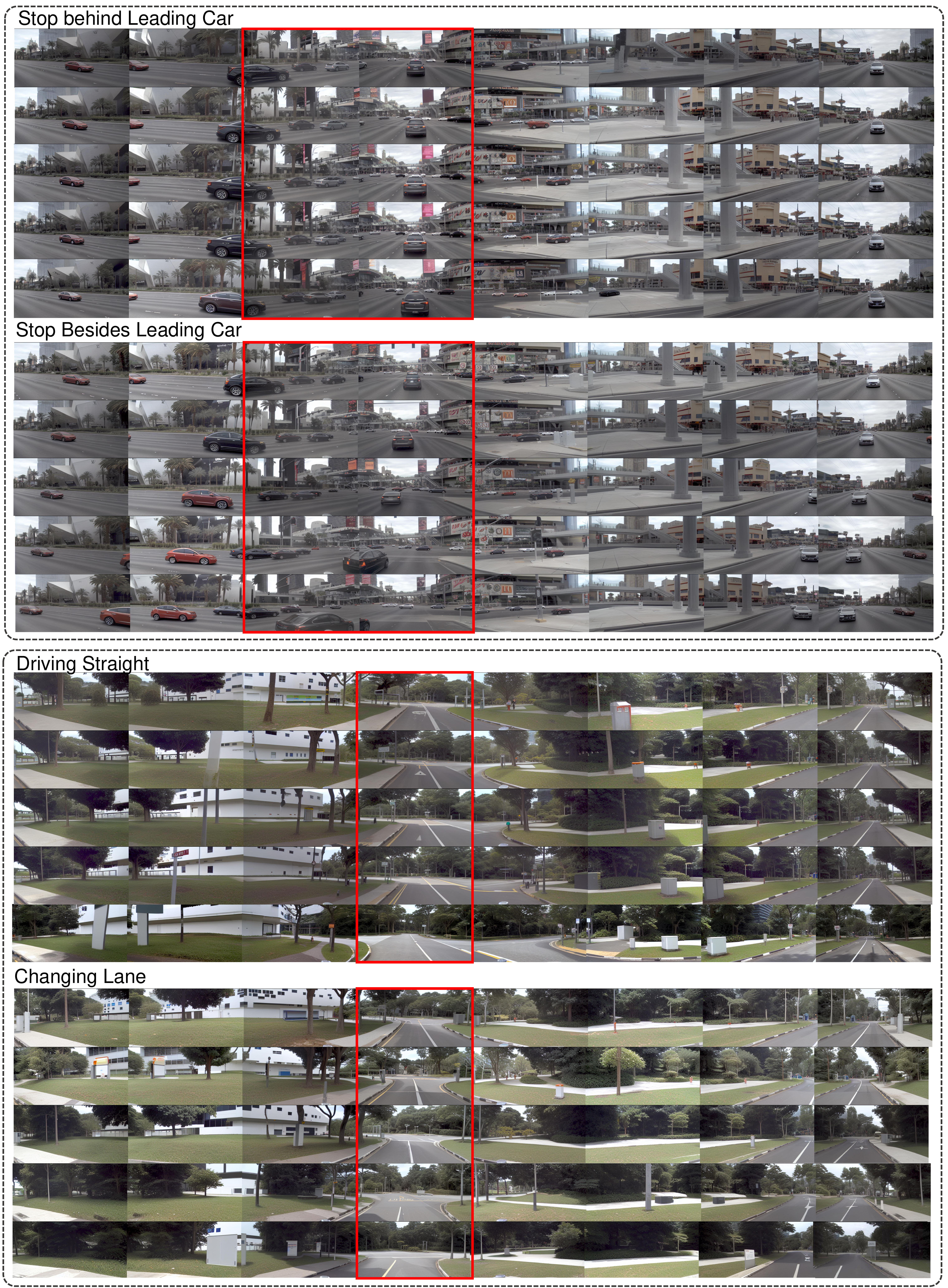}
    \caption{\textbf{Interactive Simulation Result}. Views with most conspicuous differences are highlighted with red boxes.}
    \label{fig:inter3}
\end{figure*}

We provide more interactive simulation results in Fig.~\ref{fig:inter1}~\ref{fig:inter2}~\ref{fig:inter3}. For each driving scenario, we let the ego driving agent conduct two different behaviors. Bench2Drive-R ensures great spatial-temporal consistency, providing a coherent simulation environment.

\section{Conclusion}
We present Bench2Drive-R, a simulation-oriented generative framework that enables reactive closed-loop evaluation for end-to-end driving models. We prove the efficacy of the simulation-oriented designs through thorough experiments.

\clearpage

{
    \small
    \bibliographystyle{ieeenat_fullname}
    \bibliography{main}
}

\end{document}